\documentclass[preprint,12pt]{elsarticle}

\usepackage{amssymb}

\usepackage{amsmath,amsfonts}
\usepackage{algorithmic}
\usepackage{algorithm}
\usepackage{array}
\usepackage[caption=false,font=normalsize,labelfont=sf,textfont=sf]{subfig}
\usepackage{textcomp}
\usepackage{stfloats}
\usepackage{url}
\usepackage{verbatim}
\usepackage{graphicx}
\usepackage{booktabs}
\usepackage{colortbl}
\usepackage{multirow}
\usepackage[table,xcdraw]{xcolor}
\usepackage{soul}
\usepackage{caption}

\usepackage{romannum}
\usepackage{makecell}

\definecolor{Gray}{gray}{0.9}
\definecolor{DarkGray}{gray}{0.8}
\definecolor{LightGray}{gray}{0.97}

\usepackage{hyperref}
\usepackage{xurl}
\usepackage{subfig}
\usepackage{subcaption}
\usepackage{newtxtext,newtxmath}  
\journal{Internet of Things}

\begin{document}

\begin{frontmatter}



\title{From Lab to Field: Real-World Evaluation of an AI-Driven Smart Video Solution to Enhance Community Safety}

\author[label1]{
    Shanle Yao}
    
\author[label2]{Babak Rahimi~Ardabili}

\author[label1]{Armin Danesh~Pazho}
\author[label1]{Ghazal Alinezhad~Noghre}
\author[label1]{Christopher Neff}
\author[label1]{Lauren Bourque}
\author[label1]{Hamed Tabkhi}

\affiliation[label1]{organization={Department of Electrical and Computer Engineering, University of North Carolina at Charlotte},
            city={Charlotte},
            state={NC},
            country={USA}}

\affiliation[label2]{organization={Department of Public Policy, University of North Carolina at Charlotte},
            city={Charlotte},
            state={NC},
            country={USA}}

\begin{abstract}

This article adopts and evaluates an AI-enabled Smart Video Solution (SVS) designed to enhance safety in the real world. The system integrates with existing infrastructure camera networks, leveraging recent advancements in AI for easy adoption. Prioritizing privacy and ethical standards, pose-based data is used for downstream AI tasks such as anomaly detection. A Cloud-based infrastructure and a mobile app are deployed, enabling real-time alerts within communities. The SVS employs innovative data representation and visualization techniques, such as the Occupancy Indicator, Statistical Anomaly Detection, Bird's Eye View, and Heatmaps, to understand pedestrian behaviors and enhance public safety. Evaluation of the SVS demonstrates its capacity to convert complex computer vision outputs into actionable insights for stakeholders, community partners, law enforcement, urban planners, and social scientists. This article presents a comprehensive real-world deployment and evaluation of the SVS, implemented in a community college environment with 16 cameras. The system integrates AI-driven visual processing, supported by statistical analysis, database management, cloud communication, and user notifications. Additionally, the article evaluates the end-to-end latency from the moment an AI algorithm detects anomalous behavior in real-time at the camera level to the time stakeholders receive a notification. The results demonstrate the system's robustness, effectively managing 16 CCTV cameras with a consistent throughput of 16.5 frames per second (FPS) over a 21-hour period and an average end-to-end latency of 26.76 seconds between anomaly detection and alert issuance.

\end{abstract}


\end{frontmatter}
%

\section{Introduction}\label{int}

Closed-Circuit Television (CCTV) cameras are widely applied in public spaces, traditionally serving as passive surveillance tools aimed primarily at security and monitoring \cite{ashby2017value}. However, these systems often underutilizing the potential of visual data they capture, missing opportunities to contribute to broader civic applications such as urban planning, public health, and resource management \cite{angelidou2018enhancing}. By transforming CCTV networks into Smart Video Solutions (SVS), we can unlock new functionalities that go beyond mere surveillance to enhance community living and environments \cite{javed2022future}.

SVS utilizes advanced AI and machine learning techniques to conduct real-time analysis of video data, thereby generating actionable insights that enhance urban infrastructure and community services \cite{ardabili2023understanding}. Specifically, these insights can inform urban planners about pedestrian flows to improve traffic management \cite{rashvand2023real}, enable public health officials to monitor social distancing \cite{usmani2023deep}, and facilitate the efficient allocation of community resources \cite{dechouniotis2020edge}. Collectively, these applications underscore the significant potential of SVS to transform video data into valuable information for civic improvements.

While previous research has focused on developing SVS in controlled environments \cite{pazho2023ancilia}, transitioning and evaluating these systems in real-world settings is crucial to address challenges such as latency, scalability, and privacy concerns \cite{intro7, intro9}. This shift from controlled laboratory settings to diverse real-world environments ensures that SVS is robust against real-world variability and adaptable to changing conditions, enabling it to effectively handle the complexities of dynamic urban environments and diverse community needs \cite{noghre2023understanding}. This study primarily focuses on evaluating of such a system in real-world settings, leveraging both the AI \cite{pazho2023ancilia} and application \cite{ardabili2023understanding} components from the earlier State-Of-The-Art (SOTA) research that emphasized the development of such a system. Through this real-world evaluation, we are assessing the system's impact in community settings and its broader civic applications.

\begin{figure*}[]
        \centering
               \includegraphics[width=1\linewidth, trim= 20 600 20 10,clip]{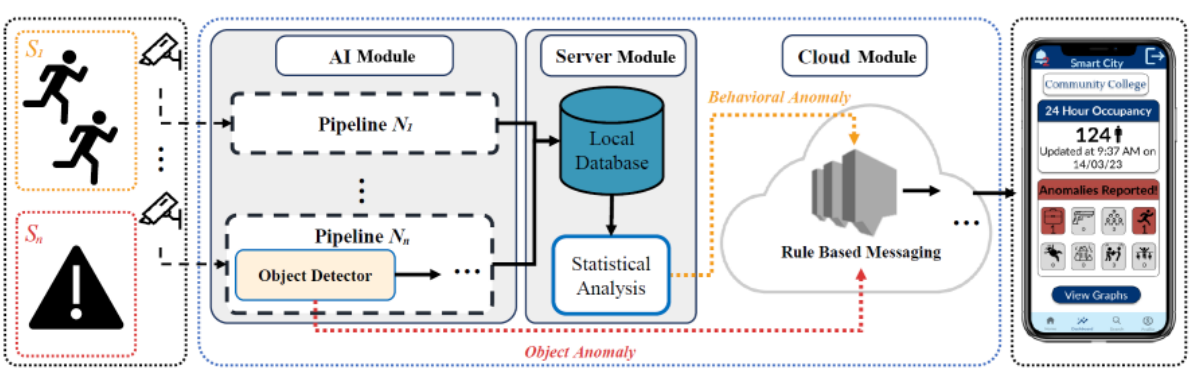}
               \caption{The end-to-end system architecture is coupled with the notification's data flow. \textit{$S_{1}$} represents the first scene that detects a behavior anomaly. The yellow line shows the notification data flow to the notification service. \textit{$S_{n}$} shows the \textit{$n^{\text{th}}$} scene where a suspicious object has been detected. The red line shows the notification data flow when detecting object anomalies.}       
                \label{fig:Concept}
\end{figure*}

The architecture of the SVS is illustrated in Fig. \ref{fig:Concept}, which represents the end-to-end design \cite{ardabili2023understanding}. It begins with the environment where video data is captured by existing CCTV cameras. The data is then processed by the AI module, where computer vision techniques are applied for human behavior analysis. The processed information is sent to the server module for further data analysis and aggregation \cite{pazho2023ancilia}. To ensure privacy, metadata is stored in the cloud module, which also hosts a mobile application. This setup allows end-users to receive notifications and access insights on their devices, fostering real-time engagement and decision-making.

Previous studies have mainly focused on the development of SVS either from the technical or the application perspectives. However, their approaches were mainly limited to lab settings, which lacked real-world validations. Deploying these systems in the real world, introduce new challenges such as network latency, integration with the existing infrastructure, and scalability. Moreover, the main scope of SVS focuses on security applications like detecting behavior anomalies or detecting fire hazards. Generally, there is less debate on privacy in these  applications, where the focus is on immediate identification of security threats. However, when there is a need for continuous monitoring of public spaces in broader civic applications,  privacy concerns become significantly more important. This further underscores the need to assess real-world performance to ensure that the system operates effectively while complying with ethical aspects. In this study, we bridge the gap in deploying and evaluating an SVS in the real-world setup and go beyond security applications, to show the scalability and robustness in broader civic applications. This article introduces the deployment and evaluation of an AI-enabled SVS designed to enhance safety and civic functionalities in community spaces such as educational areas, parking lots, and smart city applications. By employing privacy-preserving techniques, the system utilizes pose-based data for human behavior analysis, aligning with ethical standards. The SVS applies innovative data visualization tools, including Occupancy Indicators, Statistical Anomaly Detection, Bird's Eye View, and Heatmaps, to provide insights into pedestrian behaviors and resource allocation, thereby aiding urban planning and enhancing public safety.

The SVS has been implemented in a community college to evaluate its real-world effectiveness. Our findings demonstrate the system's capability to manage 16 CCTV cameras with a throughput of 16.5 frames per second (FPS) over a 21-hour period, maintaining an average latency of 26.76 seconds for detecting behavioral anomalies and notifying users.

In summary, the contributions of this article are:

\begin{itemize}
    \item Adopting an AI-enabled SVS with existing CCTV infrastructure in the real world, emphasizing civic applications beyond traditional security measures.
    \item Presenting innovative data representation techniques that enhance urban planning and resource allocation by providing insights into pedestrian behaviors.
    \item Providing a comprehensive real-world evaluation of the SVS, demonstrating its effectiveness, scalability, and privacy-preserving capabilities in community settings.
\end{itemize}

The remainder of this paper is structured as follows. Section \ref{sec:related-works} provides a comprehensive review of relevant literature in the field. Section \ref{sec:System-Features} outlines the key features and architecture of our system. In Section \ref{sec:Test_Setup}, we detail the implementation methodology and evaluation framework. Section \ref{sec:Data} presents our data visualization tools and techniques, while Section \ref{sec:engagement} examines community engagement and its impact. Section \ref{sec:Results} describes our experimental setup and discusses the results. Finally, Section \ref{sec:Conclusion} summarizes our findings and discusses future research directions.

\section{Related Works}\label{sec:related-works}

Recent advancements in AI-driven video systems have harnessed state-of-the-art algorithms and system designs to execute high-level AI tasks through computer vision. For example, Ancilia \cite{pazho2023ancilia} utilizes a multi-stage computer vision pipeline specifically engineered to enhance public safety. Although we adopted this system as the baseline for our deployment, the original study reports results exclusively from a controlled environment. This limitation underscores the necessity of real-world evaluations to confirm the system’s practical efficacy.

Deploying SVS systems in real-world environments is critical to validate their performance and tackle key challenges, including latency, scalability, and privacy concerns. Historically, research has concentrated on laboratory-based studies for real-time object tracking and anomaly detection \cite{pazho2222, RL2}. However, there is increasing acknowledgment of the need to assess these systems in dynamic, real-world settings. Such evaluations are vital to ensure SVS technologies can adapt to the complexities of urban landscapes and fulfill diverse community requirements.

Recent advancements in machine learning have enabled smart video technology to perform complex data analysis, offering insights for civic applications. For instance, Pramanik et al. demonstrated lightweight models for edge devices to reduce latency in urban environments \cite{RL4}. Singh et al. introduced an AI-based video technology for fall detection, highlighting the potential for applications beyond traditional safety measures \cite{RL5}. 

In 2023, the E2E-VSDL method by Gandapur utilized BiGRU and CNN for anomaly detection, achieving high accuracy \cite{RL6}. Similarly, TAC-Net's deep learning approach excelled in addressing anomaly scenarios \cite{RL7}. 

Integrating these systems into smart city applications bridges the gap between theoretical research and practical implementation. Alshammari et al. developed an SVS using surveillance cameras in real-world settings \cite{RL8}, and RV College of Engineering Bengaluru improved object detector accuracy \cite{RL9}. These efforts highlight the necessity of robust testbed support for evaluating SVS, addressing scalability and privacy challenges \cite{RL10}. Y. Yuan et al. \cite{RL12} introduced an innovative decentralized framework for Cyber-Physical systems, demonstrating its effectiveness in optimizing traffic flow and reducing pedestrian waiting times in real-world scenarios. While \cite{RL15} presents an impressive self-aware Cyber-Physical system, it notably lacks comprehensive data on system latency—a crucial metric in testbed environments.

Recent research has increasingly focused on optimizing large-scale Smart Vehicle systems, with significant attention given to enhancing data transformation, communication efficiency, and the integration of blockchain technology across various domains \cite{RL11}. These developments demonstrate a shift beyond traditional model performance metrics to address comprehensive system scalability and real-world implementation challenges.

The combination of machine learning and IoT devices has transformed urban planning, providing insights for managing congestion and enhancing civic participation. Ardabili et al. introduced an end-to-end system design that integrates Ancilia with an end-user device for direct communication. We have adopted this comprehensive system design for our study's deployment, allowing us to assess its effectiveness in practical, real-world environments \cite{ardabili2023understanding}. Mahdavinejad et al. emphasized machine learning's role in forecasting urban congestion \cite{mahdavinejad2018machine}, while Zanella et al. advocated for accessible urban IoT data \cite{zanella2014internet}. These studies illustrate how data-driven approaches contribute to civic applications.

Despite significant advances in Smart Vehicle Systems research, there remains a critical gap in the comprehensive testbed evaluation of real-world SVS performance. To address this limitation, we have developed and implemented an SVS system within a public community college setting, providing a practical demonstration of these technologies in a real-world environment. The following sections detail our implementation, findings, and their implications for future SVS deployments.

\section{Software System Features}\label{sec:System-Features}

 \begin{figure*}[]
        \centering
               \includegraphics[width=1\linewidth, trim= 10 290 10 30,clip]{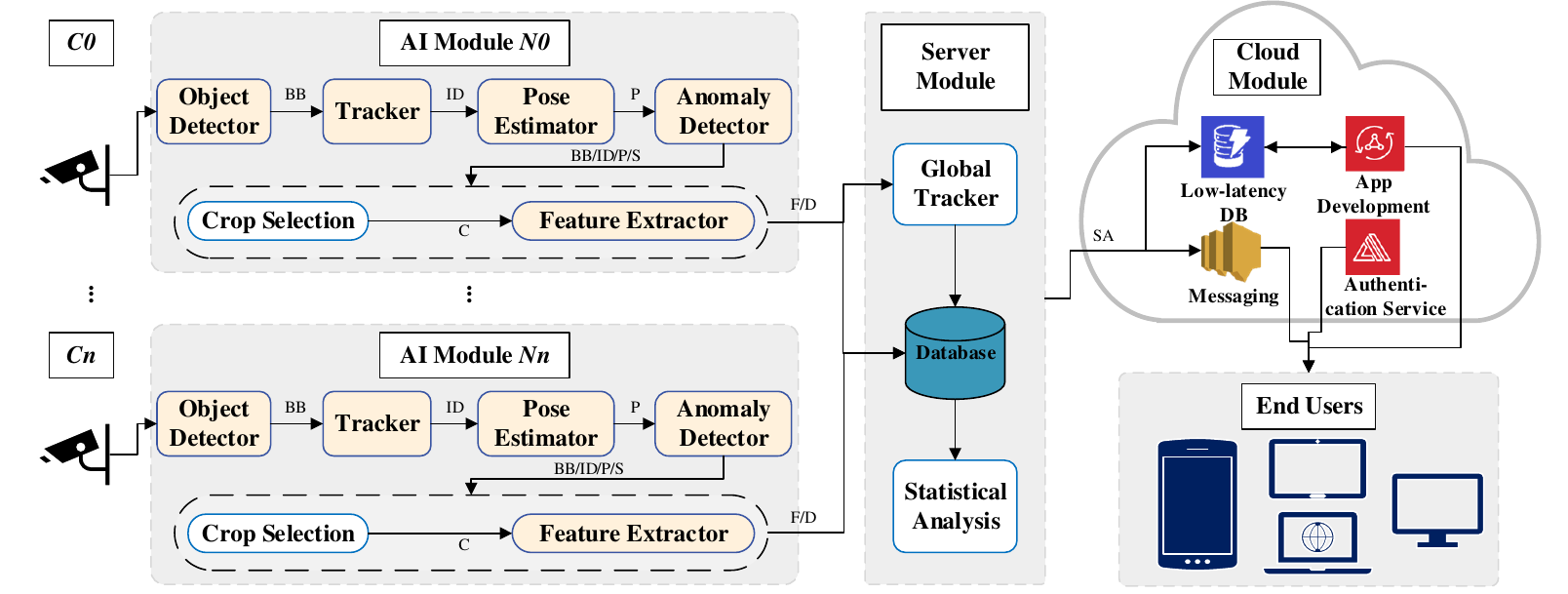}
                    \caption{End-to-end detailed system. \textit{$C_{0}$} represents the camera \textit{ID} and for each camera, one AI Module \textit{$N_{0}$} including a multi-AI-vision-models pipeline is assigned. All the AI Modules send processed data to one Server Module database, and the Server Module will re-identify the human track ID based on the feature extractor data. The statistical analyzer analyzes all the data stored in the database across all the cameras} and communicates the results with the Cloud Module. Cloud-native services are utilized to host the end users' applications.
                    
                \label{fig:pipe}
\end{figure*}

\begin{algorithm}
\caption{SVS Algorithm}
\label{alg:system}
\begin{algorithmic}[1]

\STATE \textbf{Initialization:} 
\STATE \quad Initialize AI modules, Server Module, and Cloud Module.

\FOR{\textbf{each} camera $c_i$ \textbf{in} the set of cameras}
  \WHILE{\textit{stream from} $c_i$ \textit{is active}}
  
    \STATE \textbf{Step 1: Object Detection}
    \STATE \quad Fetch a batch of 30 frames of $rgb$ data from $c_i$ as \textbf{initial input}
    \STATE \quad Apply the object detector to identify humans and objects of interest.
    \STATE \quad Record bounding boxes $B$ with associated confidence scores.
    \IF{\textbf{objects of interest} confidence score in $B \geq \textit{object\_threshold}$}
      \STATE Trigger an alert for the detected object of interest to end-user device(s).
    \ELSE
      \STATE {Continue}
    \ENDIF

    \STATE \textbf{Step 2: Tracking}
    \STATE \quad Input $B$ to the tracker to assign unique IDs and generate tracklets $T$.

    \STATE \textbf{Step 3: Pose Estimation}
    \STATE \quad For each tracklet in $T$, extract the corresponding input $rgb$ data.
    \STATE \quad Apply the pose estimator to compute skeletal keypoints $K$.

    \STATE \textbf{Step 4: Anomaly Score Prediction}
    \STATE \quad Organize input keypoints into movement sequences.
    \STATE \quad Compute anomaly scores $\alpha$ for the batch of keypoints.

    \STATE \textbf{Step 5: Feature Extraction \& Cross-Camera Identification}
    \STATE \quad Extract feature $F$ from input $rgb$ for cross-camera identification.
    \STATE \quad Transmit $F$ and $\alpha$ to the Server Module for global analyze.

    \STATE \textbf{Step 6: Behavior Anomaly Detection}
    \STATE \quad Transmit $\alpha$ to the Server Module for behavior anomaly detection.
    \IF{$\alpha \geq \textit{threshold}$ \textbf{for three consecutive batches}}
      \STATE Trigger an alert to the end-user device(s).
    \ELSE
      \STATE {Continue}
    \ENDIF

    \STATE \textbf{Step 7: Data Processing at Server Module}
    \STATE \quad Process $F$, anomaly scores $\alpha$, and related data for statistical analysis.

    \STATE \textbf{Step 8: Data Transmission}
    \STATE \quad Transmit analyzed results to the Cloud and output results to end-user device(s) with different visualizations.
    
  \ENDWHILE
\ENDFOR

\end{algorithmic}
\end{algorithm}

The AI-based real-time video solution seamlessly integrates with existing CCTV infrastructures, creating a Physical-Cyber-Physical (PCP) system that delivers actionable information to end users. Such a system consists of four key components: AI Modules, Server Module, Cloud Module, and End-user Devices, as illustrated in Fig. \ref{fig:pipe}. Multiple camera streams can be fed into separate AI Modules in parallel. The object detector identifies bounding boxes for humans and other objects of interest, then forwards this bounding-box data to a tracker, which assigns unique IDs. Next, the tracked bounding boxes, including associated pixel data, are passed to a pose estimator to determine key skeletal joints. These skeletal data joints, organized into movement sequences, are subsequently processed by an anomaly detector to produce an anomaly score for the scene. All previously processed data are then sent to a feature-extraction stage to generate unique signatures for cross-camera identification, a task carried out by the server module. Finally, additional statistical data are analyzed and transmitted to the cloud module for end-user applications. The design of the AI and server modules was inspired by prior work \cite{pazho2023ancilia}, while the cloud module and user device components were adapted from another study \cite{ardabili2023understanding}. Algorithm \ref{alg:system} provides more details regarding the complete SVS.

\subsection{AI Module/Modules}

AI Module/Modules is a multi-stage pipeline optimized for real-time computer vision tasks. It processes image data in batches of 30 frames, in which the object detector identifies objects and the anomaly detector flags the behavior \cite{pazho2023ancilia}. Object anomaly detection and behavioral anomaly detection \cite{noghre1111, pazho1111} are performed within this module respectively, with alerts sent to end-user devices for real-time response \cite{pazho2222,Yao_2024_CVPR}.

One fundamental experiment was conducted to identify the optimal algorithm configuration for the AI node \cite{pazho2023ancilia}, specifically tailored to address real-world security requirements before going into further research. As highlighted by Pazho et al. \cite{pazho2023ancilia}, the GEPC method \cite{gepc} is particularly effective for behavioral anomaly detection. ByteTrack \cite{bytetrack} and OSNet \cite{OSnet} excel in pedestrian tracking and human re-identification, while the recent YOLOv8 \cite{yolov8} demonstrates superior performance in object detection and enhances human pose estimation through a top-down approach.

Such evaluation utilized the DukeMTMC dataset \cite{dukemtmc}, with one-minute video clips of varying crowd densities. Initial validation was conducted on a laboratory server using default weights and parameters from original repositories, measuring throughput, latency, detection outputs, and computational resource utilization. The testing protocol employed two high-density scenarios, labeled "Extreme" and "Heavy," each comprising 60-second videos at 60 frames per second, yielding 120 inference batches. Statistical validity was ensured by analyzing the middle 80 batches exclusively, eliminating potential bias from warm-up and cool-down periods in the performance metrics.

Four distinct methods underwent testing:
\begin{itemize}
\item YOLOv5+HRNet: The original system method 
\item YOLOv8pose: A top-down human pose estimation method.
\item YOLOv8pose-p6: The most complex model of YOLOv8 for pose estimation.
\item YOLOv8+HRNet: Object detection method with human pose estimation as a bottom-up method.
\end{itemize}

\begin{table*}[]
\small  
\setlength{\tabcolsep}{1pt}  
\centering
\caption{Performance comparison among different system configurations. Data collected using lab server with a single node. }
\label{tab:model-comparison}

\begin{tabular}{c|c|c|c|c|c|c|c}
\rowcolor{DarkGray}
\begin{tabular}[c]{@{}c@{}}Crowd \\ Density \end{tabular}& Method & \begin{tabular}[c]{@{}c@{}}Latency\\ (s)\end{tabular} &

FPS & 
\begin{tabular}[c]{@{}c@{}}Detections\\ (counts)\end{tabular} & \begin{tabular}[c]{@{}c@{}}GPU \\ Memory\\ (G)\end{tabular} & \begin{tabular}[c]{@{}c@{}}CPU \\ Memory\\ (G)\end{tabular} & \begin{tabular}[c]{@{}c@{}}Total \\ FLOPs\\ (B)\end{tabular} \\ \hline
\multirow{4}{*}{\begin{tabular}[c]{@{}c@{}}Extreme\\ ($\sim$50 detects \\ per second)\end{tabular}} & YOLOv5+HRNet & 14 & 14.82 & 1033 & \textbf{5.3} & 19.1 & \textbf{126.08} \\
 & YOLOv8pose-p6 & 3.7 & 26.39 & 911 & 12.2 & 12.68 & 1067.38 \\
 & YOLOv8pose & \textbf{2.2} & \textbf{49.07} & 679 & 6.9 & \textbf{12.38} & 264.18 \\
 & YOLOv8+HRNet & \textit{9.09} & \textit{19} & \textit{\textbf{1217}} & \textit{5.5} & \textit{19.3} & \textit{182.18} \\ \hline
\multirow{4}{*}{\begin{tabular}[c]{@{}c@{}}Heavy\\ ($\sim$20 detects \\ per second)\end{tabular}} & YOLOv5+HRNet & 3.47 & \textbf{81.2} & 237 & \textbf{5.0} & 15.28 & \textbf{126.08} \\
 & YOLOv8pose-p6 & 3.54 & 26.45 & \textbf{358} & 11.9 & 12.4 & 1067.38 \\
 & YOLOv8pose & \textbf{2.6} & 45.11 & 308 & 6.9 & \textbf{12.3} & 264.18 \\
 & YOLOv8+HRNet & \textit{2.76} & \textit{67.63} & \textit{342} & \textit{5.5} & \textit{15.4} & \textit{182.18}
\end{tabular}
\end{table*}

Table \ref{tab:model-comparison} evaluates system performance across varying crowd densities using four key metrics: average latency and throughput, total detection count, average GPU and CPU memory utilization, and the AI models' total FLOPs. The combined YOLOv8+HRNet architecture demonstrated superior detection capabilities, particularly in challenging "Extreme" and "Heavy" crowd density scenarios, validating its robustness for real-world applications. For a comprehensive evaluation, our pipeline integrates YOLOv8 \cite{yolov8} for object detection, ByteTrack \cite{bytetrack} for tracking, HRNet \cite{hrnet} for pose estimation, GEPC \cite{markovitz2020graph} for behavioral analysis, and OSNet \cite{OSnet} for re-identification. 

\begin{table}[]
\centering
\small
\caption{Accuray of different model weights implemented in the system}
\begin{tabular}{c|c|cc}
\rowcolor{DarkGray}
\textbf{Model} & \textbf{Dataset} & \textbf{Metric} & \textbf{Value} \\ \hline
YoloV8l \cite{yolov8} & COCO & mAP & 52.9 \\ \hline
ByteTrack \cite{bytetrack} & MOT17 & MOTA & 80.3 \\ \hline
HRNet\_w32 \cite{hrnet} & COCO & AP & 0.744 \\ \hline
GEPC \cite{gepc} & NTU60 & AUC-ROC & 0.85 \\ \hline
OSNet \cite{OSnet} & OSNet & R1 & 94.8
\end{tabular}
\label{tab:accuracy}
\end{table}

Importantly, all accuracy metrics in subsequent experiments are derived directly from the pre-trained model weights as published in each respective repository, without any additional fine-tuning or optimization. This approach ensures reproducibility and maintains consistency with the established benchmarks reported in the original publications. Table \ref{tab:accuracy} summarizes the accuracy metrics of each model that is used for experiments on the existing well-established datasets.

\subsection{Server Module}

\begin{table}[htbp]
\small  
\setlength{\tabcolsep}{3pt}  
\centering
\caption{Example visualization of the database at Server Module}
\label{tab:database}
\begin{tabular}{c|c|c|c|c|c|c|c}
\rowcolor{DarkGray}
\textbf{\makecell{Record\\Time}} & 
\textbf{Camera} & 
\textbf{Class ID} & 
\textbf{\makecell{Bounding\\Box}} & 
\textbf{Feature} & 
\textbf{\makecell{Local\\ID}} & 
\textbf{\makecell{Global\\ID}} & 
\textbf{\makecell{Anomaly\\Score}} \\ \hline
00:00:00 & 1 & 0 & [x, y, w, h] & Tensors & 15 & 1001 & 40 \\
00:00:01 & 2 & 0 & [x, y, w, h] & Tensors & 21 & 1001 & 40 \\
.... & ... & .. & ... & ... & .. & .. & .. \\
23:59:59 & 1 & 0 & [x, y, w, h] & Tensors & 9999 & 1001 & 40
\end{tabular}
\end{table}

Server Module centralizes metadata and historical data from AI modules, stored in a MySQL database as outlined in Table \ref{tab:database}. It handles global tracking and statistical analysis, using cosine similarity to re-identify individuals across cameras and analyze patterns \cite{pazho2023ancilia}. The design prioritizes privacy by avoiding raw data transmission to the Cloud Module.

\subsection{Cloud Module}

Cloud Module leverages cloud-native services for data storage, management, and API generation \cite{cloud}. It minimizes lag between anomaly detection and notification, using a rule-based messaging service to send real-time alerts via email, text, or app notifications. A low-latency database supports real-time data access, while APIs generated through an application development kit optimize data retrieval \cite{cloud2, ardabili2023understanding}.

\subsection{End User Devices}

End User Devices are designed to promptly notify users of detected anomalies via a smartphone application \cite{intro7}. The app provides real-time data and analysis, ensuring consistent functionality across different devices and operating systems, thereby enhancing accessibility and delivering information efficiently.

\section{Deployment and Setup}\label{sec:Test_Setup}

\begin{figure}[]
        \centering
               \includegraphics[width=0.8\linewidth, trim= 1 1 1 1,clip]{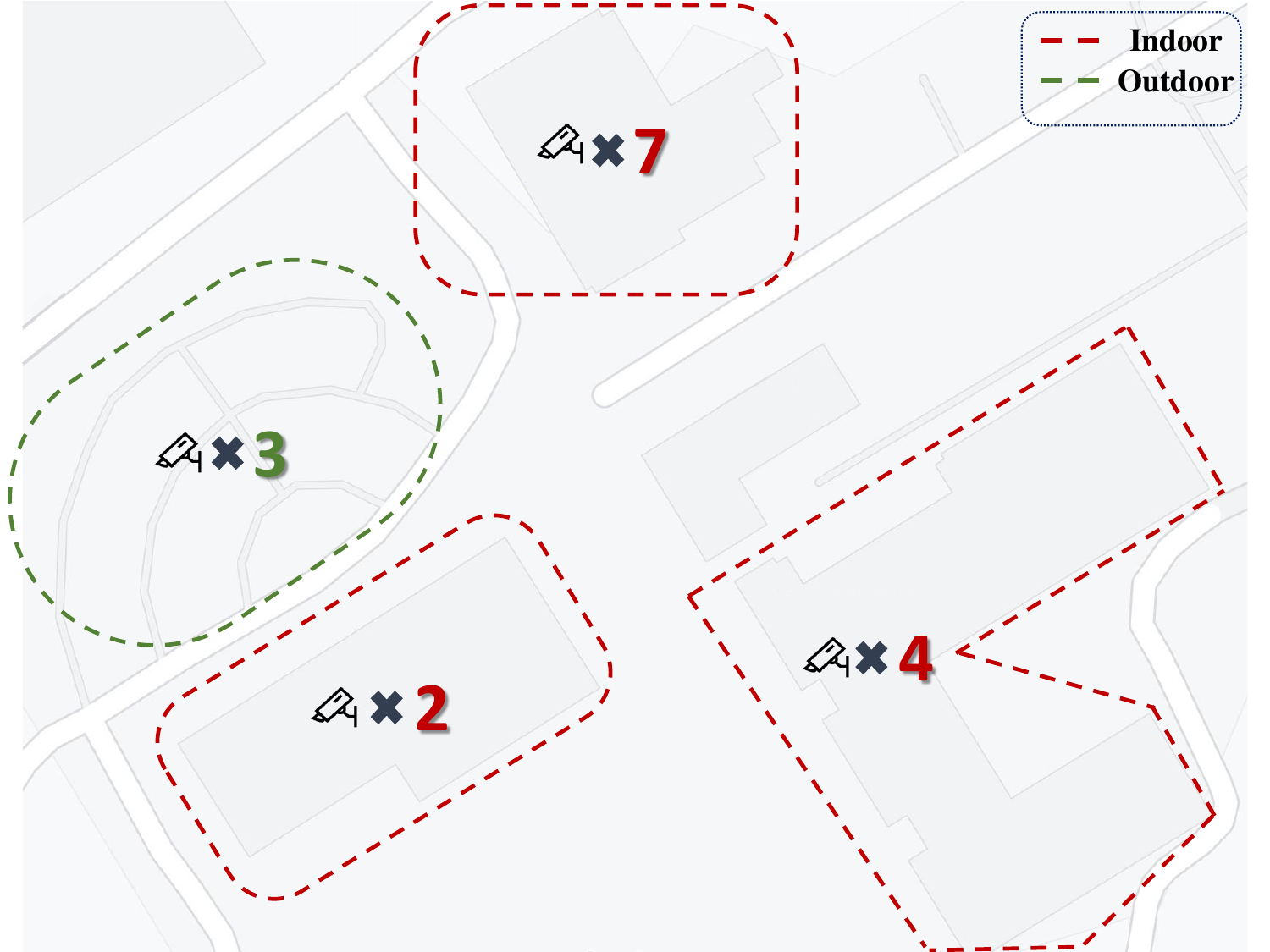}
               \caption{Locations of Cameras on the Campus. A maximum of 16 cameras are used}
                \label{fig:Cam}
\end{figure}

In lab settings, each advanced software feature demonstrates significant potential individually; however, to fully realize its benefits, a comprehensive end-to-end evaluation in a real-world environment is essential. Such an evaluation allows for assessing the system's load handling and endurance under actual conditions, ensuring that it can perform reliably and effectively when deployed, which might not be evident in a controlled environment. 

Our study utilizes an existing surveillance network of 16 AXIS IP cameras, operating at 30 FPS with 720p resolution, across the campus, monitoring approximately 95,000 square feet of combined indoor and outdoor space. As shown in Fig. \ref{fig:Cam}, thirteen indoor cameras are mounted and angled slightly downward to capture hallways, entryways, vending areas, and other common spaces, while three outdoor cameras, installed at 10 ft 8 in, oversee building exteriors, walkways, and surrounding grounds. All cameras feature varifocal lenses offering a 36--100° horizontal field of view, allowing flexible adjustments to accommodate wide-area observation. The system runs on a dedicated server with a 16-core CPU, 252 GB RAM, and four 24 GB VRAM GPUs.

The software components, encompassing the AI module and the Server Module as introduced in Section \ref{sec:System-Features}, are integrated into a dedicated server infrastructure. This server is interconnected within the local network of the testbed to enhance security measures. Video feeds from all cameras are transmitted utilizing RTSP. Data processed by each AI module is subsequently relayed to the database housed within the Server Module via TCP. The Cloud module server periodically aggregates data from the Server Module’s database for advanced data management purposes. The resultant outputs are then disseminated and updated across end-user applications

\begin{figure}[]
        \centering
               \includegraphics[width=0.8\linewidth, trim= 600 370 300 510,clip]{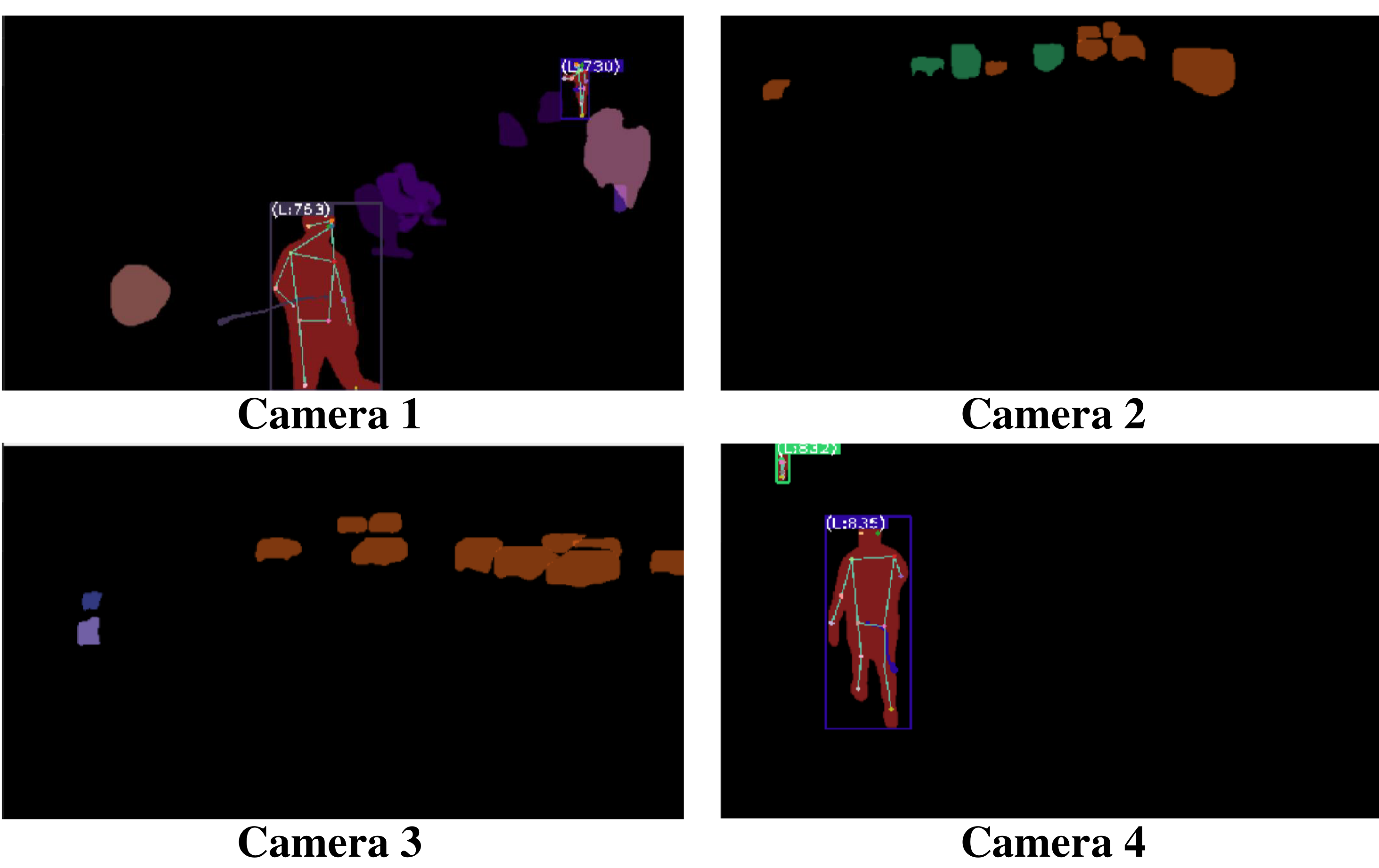}
                    \caption{This represents an example of four pipelines running in the testbed, with IP cameras deployed at different locations.}
                    
                \label{fig:testbed}
                \vspace{-10pt}
\end{figure}

As shown in Fig. \ref{fig:testbed}, a sample visualized output from the AI module during a scenario with four operational cameras is presented, with privacy-preserving instance segmentation applied.

\section{Applications and Visualizations} \label{sec:Data}

This section illustrates how the data collected from the real world can be analyzed and visualized to enhance situational awareness in different civic applications. For example, visualizing pedestrian flow can inform the design of public spaces like intersection turnarounds to reduce congestion, while tracking real-time movement patterns helps prevent overcrowding and ensures safety in public events. These applications showcase how data-driven insights can lead to more efficient, safe, and well-planned communities.

\subsection{Descriptive Data}\label{Desc.}

In our descriptive analysis, the Global ID from human feature data serves as the unique identifier for tracking across data streams. We present five key metrics to provide a general overview of traffic flow and occupancy trends: real-time count of people, hourly average number of people per camera per location, total number of people, and peak hour analysis over time. These metrics help monitor occupancy, understand distribution trends, and identify peak traffic times, aiding in resource allocation and emergency planning.

\subsection{Situational Awareness}\label{Aware}

Situational awareness is crucial for effective decision-making across various domains. Our study explores the four following key visualization techniques:

\subsubsection{Occupancy Indicator}\label{OI}

The Occupancy Indicator provides insight into the number of people in a specific location by comparing current occupancy levels to historical data, which is particularly useful in emergencies or public health scenarios. To calculate occupancy, the system compares current detections with historical percentiles, categorizing occupancy as "Low," "Normal," or "High." 

\begin{figure}[]
        \centering
               \includegraphics[width=0.9\linewidth,trim= 0 150 0 0,clip]{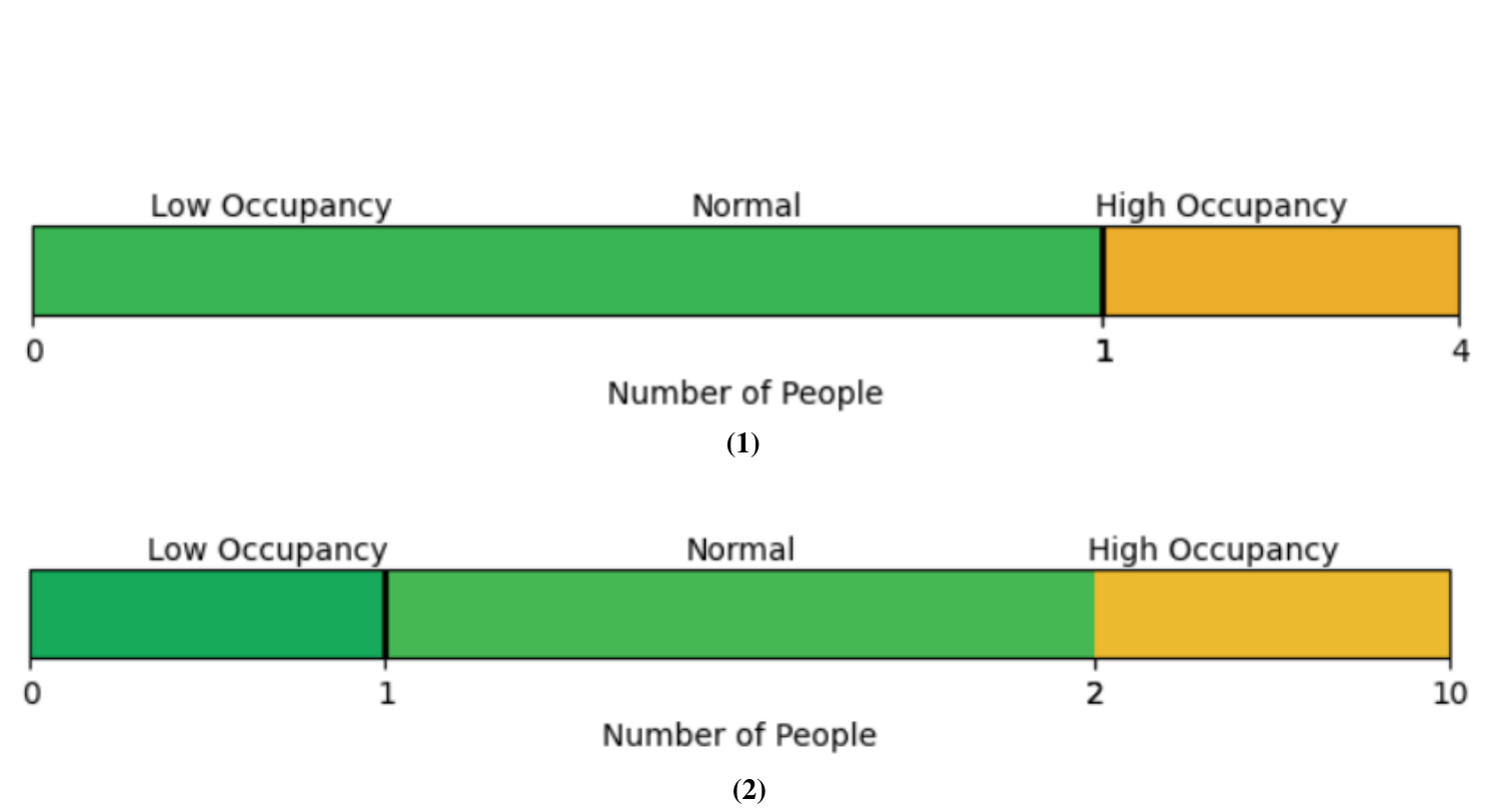}
                    \caption{Comparing the Occupancy Indicator of the same camera during the same hour on two different days}. Graph (1) shows the Weekend. Graph (2) represents weekday.
                \label{fig:OI}
\end{figure}

Algorithm \ref{alg:occupancy} describes a method for calculating and categorizing real-time occupancy levels across multiple camera locations. Operating on a continuously updated data frame (\texttt{df}), the algorithm processes each camera location sequentially. For each camera ID, it performs occupancy analysis at 5-second intervals throughout the specified \texttt{record time}. During each interval, the algorithm computes the current occupancy by determining the number of unique \texttt{global IDs} present in the data frame, effectively tracking the distinct individuals within each camera's field of view.

\begin{algorithm}
   \caption{Occupancy Indicator Algorithm}
   \label{alg:occupancy}
\begin{algorithmic}[1]
   \STATE \textbf{Input:} data frame $df$, historical\_data
   \FOR{each camera\_id in $df$}
       \FOR{every 5 seconds in record\_time}
           \STATE current\_\#\_of\_people = length(unique(global\_IDs))
       \ENDFOR
       \IF{current\_\#\_of\_people $\leq$ percentile(historical\_data, 25)}
           \STATE occupancy = "Low Occupancy"
       \ELSIF{current\_\#\_of\_people $\leq$ percentile(historical\_data, 75)}
           \STATE occupancy = "Normal Occupancy"
       \ELSE
           \STATE occupancy = "High Occupancy"
       \ENDIF
       \STATE update\_historical\_data(current\_\#\_of\_people)
   \ENDFOR
\end{algorithmic}
\end{algorithm}

Occupancy levels are determined by comparing the \texttt{current number of people} against historical data percentiles. The algorithm classifies occupancy into three categories based on these thresholds: 'Low Occupancy' when the count falls at or below the 25$^{th}$ percentile, 'Normal Occupancy' when it is between the 25$^{th}$ and 75$^{th}$ percentiles, and 'High Occupancy' when it exceeds the 75$^{th}$ percentile of historical data.

Fig. \ref{fig:OI} compares occupancy indicators for the same camera during the same hour on two days. Graph 1 shows a weekend, with 2 people categorized as "High Occupancy," while Graph 2 shows the same number considered "Normal" on a weekday.

Following each occupancy determination, the \texttt{update historical data} function incorporates the \texttt{current number of people} into the historical dataset. This continuous updating mechanism ensures the algorithm's thresholds remain adaptive and relevant, reflecting recent occupancy patterns. The entire process executes across all camera IDs in the data frame, providing comprehensive occupancy monitoring across the entire surveillance network.

\subsubsection{Statistical Anomaly}\label{SA}
Statistical Anomalies often indicate unusual traffic flow. By analyzing historical data, the system establishes a baseline, identifying deviations as anomalies. The system calculates such anomalies by comparing current detections with historical data, flagging events that exceed two standard deviations from the mean. In our analysis, zero detections were excluded to avoid skewing the data. 

Algorithm \ref{alg:SA} identifies statistically anomalous numbers of detected individuals for each camera relative to historical trends. Given that crowd densities vary across locations and times, these anomalies are calculated on an hourly basis. The approach involves constructing a normal distribution from historical hourly data, updated every 5 seconds as new detections occur. As outlined in Algorithm \ref{alg:SA}, the mean and standard deviation for each hour are calculated to characterize the data distribution. Current detections are then compared against this historical context. If the current number of detections exceeds two standard deviations from the mean, there is a less than 0.05 probability of such an occurrence, indicating with 95\% confidence that it can be considered a statistical anomaly.

\begin{algorithm}[tb]
   \caption{Statistical Anomaly Detection}
   \label{alg:SA}
\begin{algorithmic}[1]
    \STATE \textbf{Input:} $df$, $start\_time$, $end\_time$
    \STATE Filter $df$ for camera ID in the specified time range
    \STATE Initialize $mean$ to 0 and $std$ to 1
    \STATE Initialize $detected\_objects$ list
    \FOR{each 5-second interval from $start\_time$ to $end\_time$}
        \STATE Count unique detected objects in the interval
        \IF{detected objects $>$ 0}
            \STATE Update $detected\_objects$, $mean$, and $std$
        \ENDIF
    \ENDFOR
\end{algorithmic}
\end{algorithm}

We intentionally exclude detections of \texttt{0} in our analysis. The reason is that, within a 5-second window, the absence of detections (i.e., detecting no individuals) is more likely than detecting any individuals. Including \texttt{0} detections would skew the results toward zero, leading to lower thresholds for identifying anomalies and reducing the effectiveness of the statistical analysis.

\subsubsection{Bird's Eye View}\label{BEV}

Bird's Eye View provides an accurate spatial representation, ideal for crowd management and area planning. The system normalizes object dimensions and uses a scale factor to accurately position objects within this view. 

Algorithm \ref{alg:birdseye} describes the transformation of camera-detected objects into Bird's Eye View coordinates, specifically designed for the AXIS P3225-VE Mk II Network Camera\footnote{https://www.axis.com/products/axis-p3225-ve-mk-ii/support\#technical-specifications}. This transformation creates a top-down perspective of the surveillance area.

The transformation process begins with object dimension normalization, where width and height are divided by \texttt{camera width} and \texttt{camera height}, respectively. This step ensures dimensional independence from camera resolution. Subsequently, a scale factor is computed for each object based on its normalized height and the camera's angular field of view parameters (\texttt{min theta} and \texttt{max theta}), which are obtained from the manufacturer's specifications. This scaling mechanism preserves depth perception by appropriately adjusting object sizes based on their distance from the camera.

The algorithm then calculates each object's centroid by averaging its width and height dimensions, establishing a precise reference point for positional mapping. The final Bird's Eye View coordinates (\texttt{BirdsEyeX} and \texttt{BirdsEyeY}) are derived by applying the computed scale factor to these centroid coordinates, resulting in an accurate top-down spatial representation.

For temporal analysis, the algorithm implements a data filtering mechanism that processes records at 5-second intervals, tracking unique \texttt{global ID}s to determine object counts. This filtering ensures that analysis is performed only on relevant data within specified camera IDs and time frames.

\begin{algorithm}[tb]
   \caption{Bird's Eye View Transformation}
   \label{alg:birdseye}
\begin{algorithmic}[1]
   \STATE \textbf{Input:} data frame $df$, camera\_width, camera\_height, min\_teta, max\_teta
   \STATE Compute normalized values: $df['normalized\_W']$, $df['normalized\_H']$
   \FOR{each object in $df$}
       \STATE Compute $scale\_factor$ using $normalized\_H$
       \STATE Calculate centroid $C\_X$ and $C\_Y$
       \STATE Compute $BirdsEye\_X$ and $BirdsEye\_Y$ using $scale\_factor$ and normalized values
   \ENDFOR
   \FOR{$camera\_id$ from 1 to 8}
       \STATE Define start\_time and end\_time
       \STATE Filter $df$ based on $camera\_id$ and time range
   \ENDFOR
\end{algorithmic}
\end{algorithm}

\begin{figure}[t]
    \centering
    \subfloat[]
    {
        \includegraphics[width=0.43\textwidth, trim= 10 0 60 60,clip]{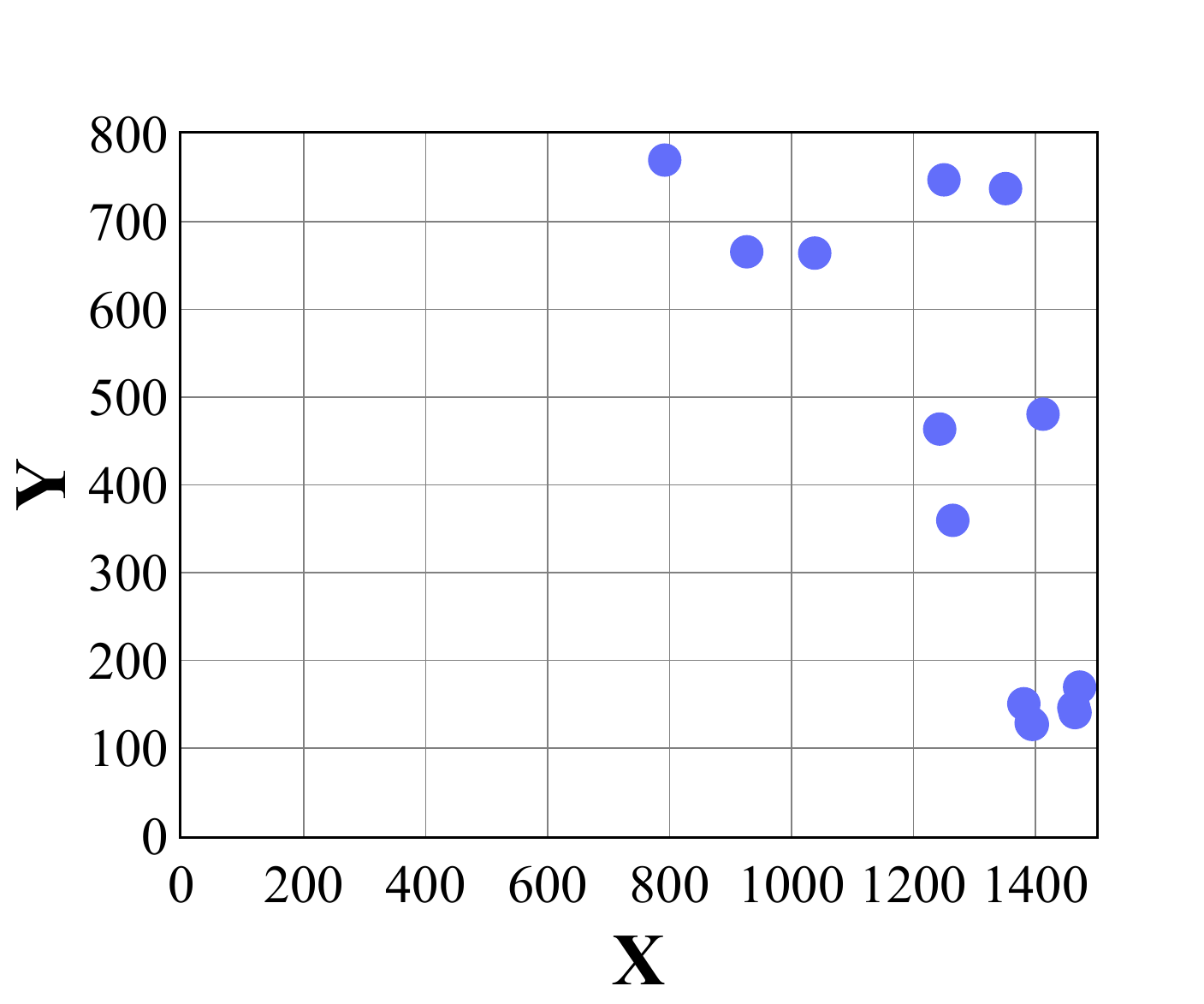}
    }
    \hfill
    \subfloat[]
    {
        \includegraphics[width=0.43\textwidth, trim= 10 0 60 60,clip]{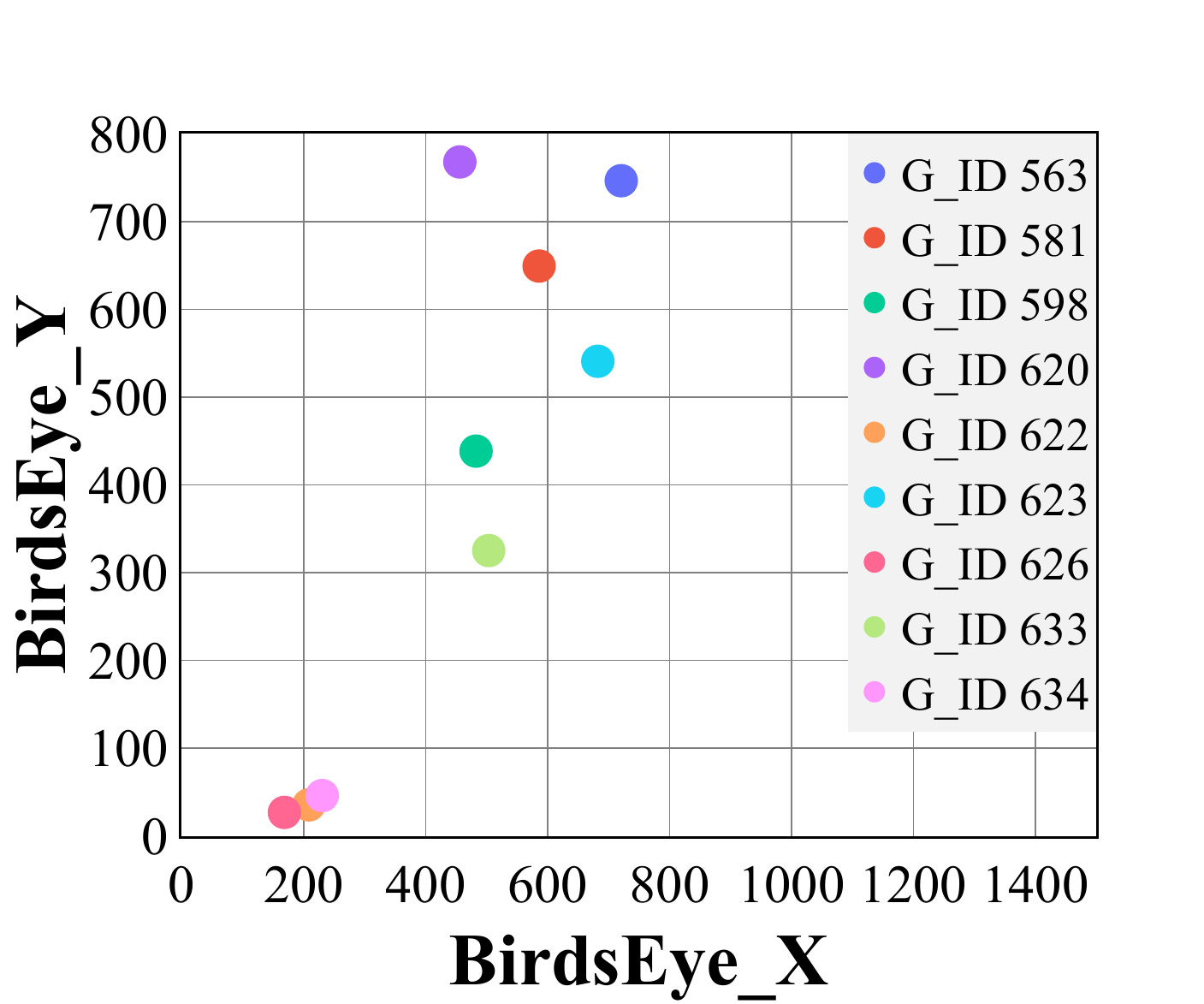}
    }
    \caption{Comparing different views from one camera. Graph (a) generated directly from the database shows the original camera view. Graph (b) represents the averaged bird's eye view coordinates.}
    \label{fig:view_graph}
\end{figure}

Fig. \ref{fig:view_graph} provides a comparative visualization between the original perspective and the processed Bird's Eye View from one camera, captured simultaneously. In the initial view, each data point represents the XY coordinates of detected objects in the camera's original field of view, showing approximately 12 distinct detections within a five-second interval. In the Bird's Eye View, these XY coordinates are transformed, consolidating objects with the same \texttt{global IDs} into averaged positions. This results in nine distinct individuals, providing a top-down perspective that refines and simplifies the spatial arrangement.

\subsubsection{Heat map}\label{Heat}

\begin{figure}[t]
    \centering
    \subfloat[Camera 2- Weekday]
    {
        \includegraphics[width=0.43\textwidth, trim= 0 0 10 60,clip]{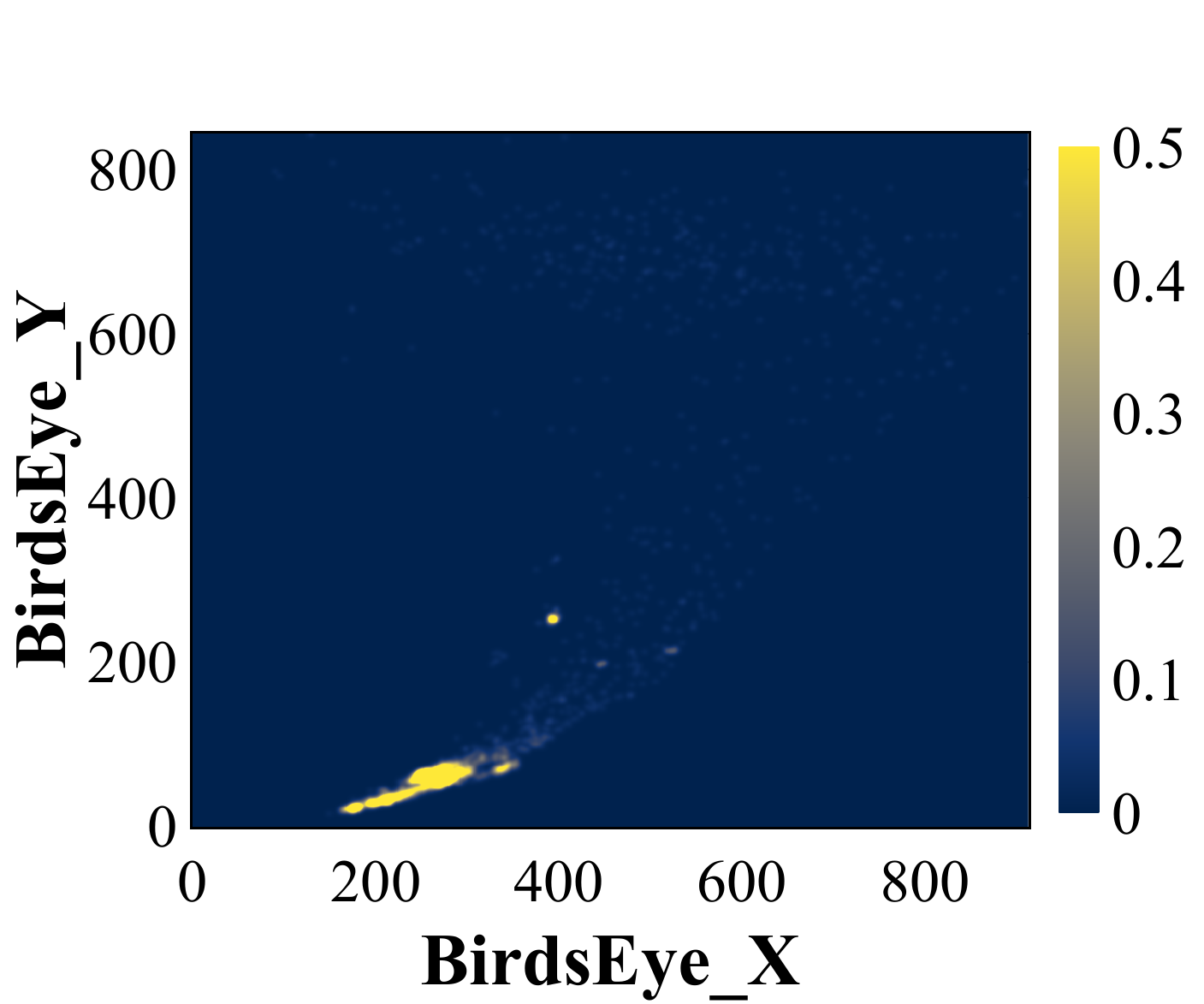}
    }
    \hfill
    \subfloat[Camera 2- Weekend]
    {
        \includegraphics[width=0.43\textwidth, trim= 0 0 10 60,clip]{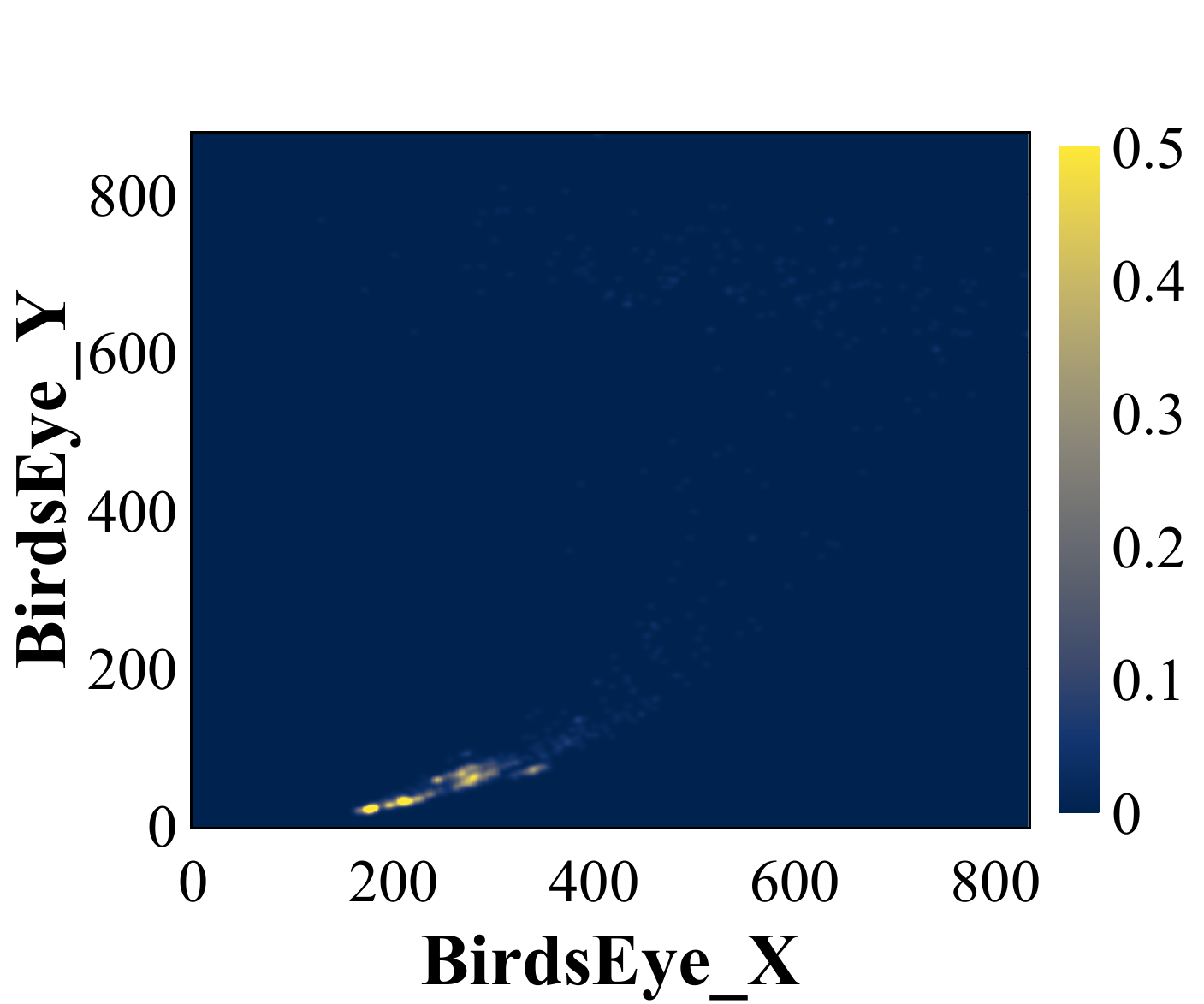}
    }
    
    \subfloat[Camera 7- Weekday]
    {
        \includegraphics[width=0.43\textwidth, trim= 0 0 10 60,clip]{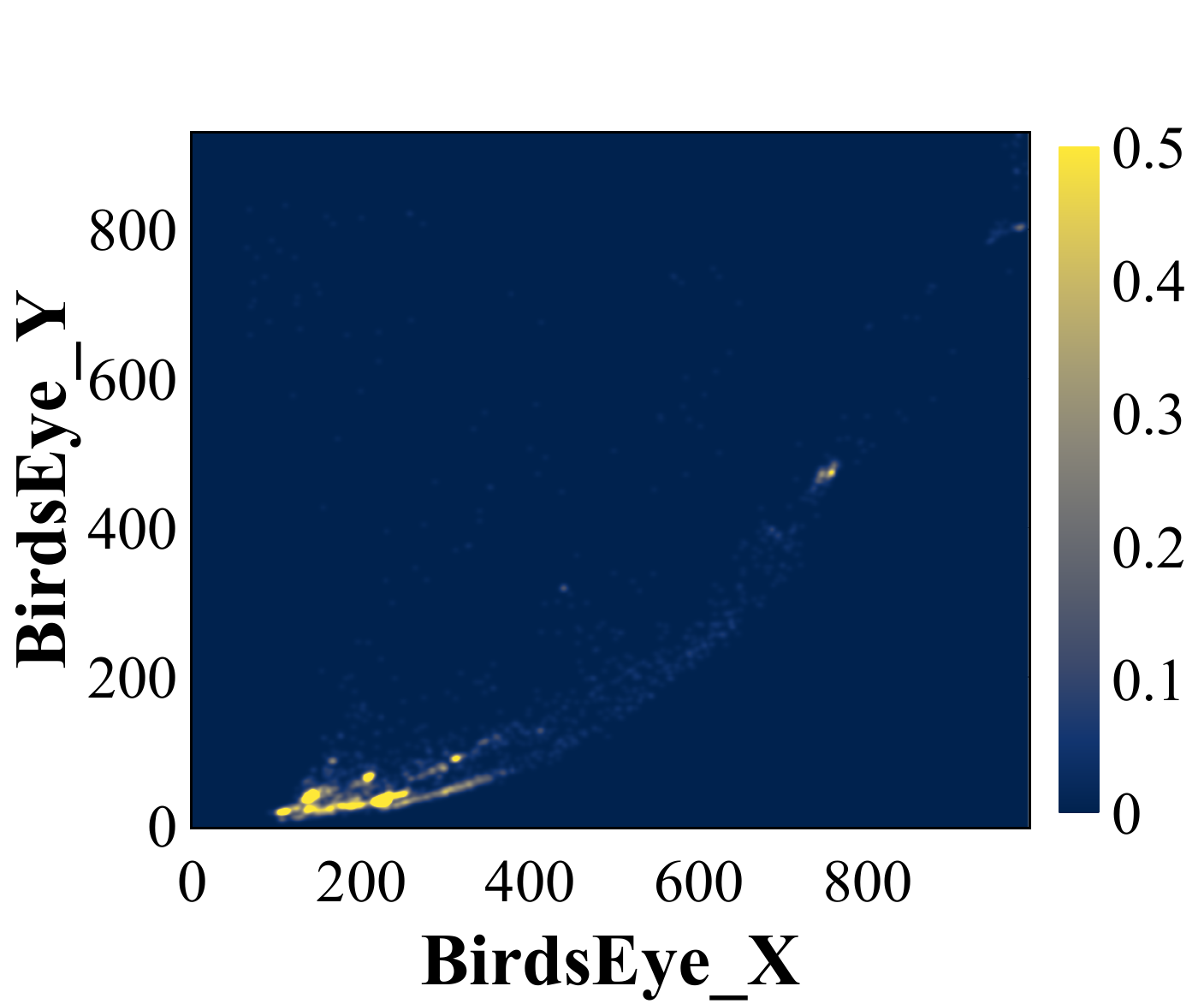}
    }
    \hfill
    \subfloat[Camera 7- Weekend]
    {
        \includegraphics[width=0.43\textwidth, trim= 0 0 10 60,clip]{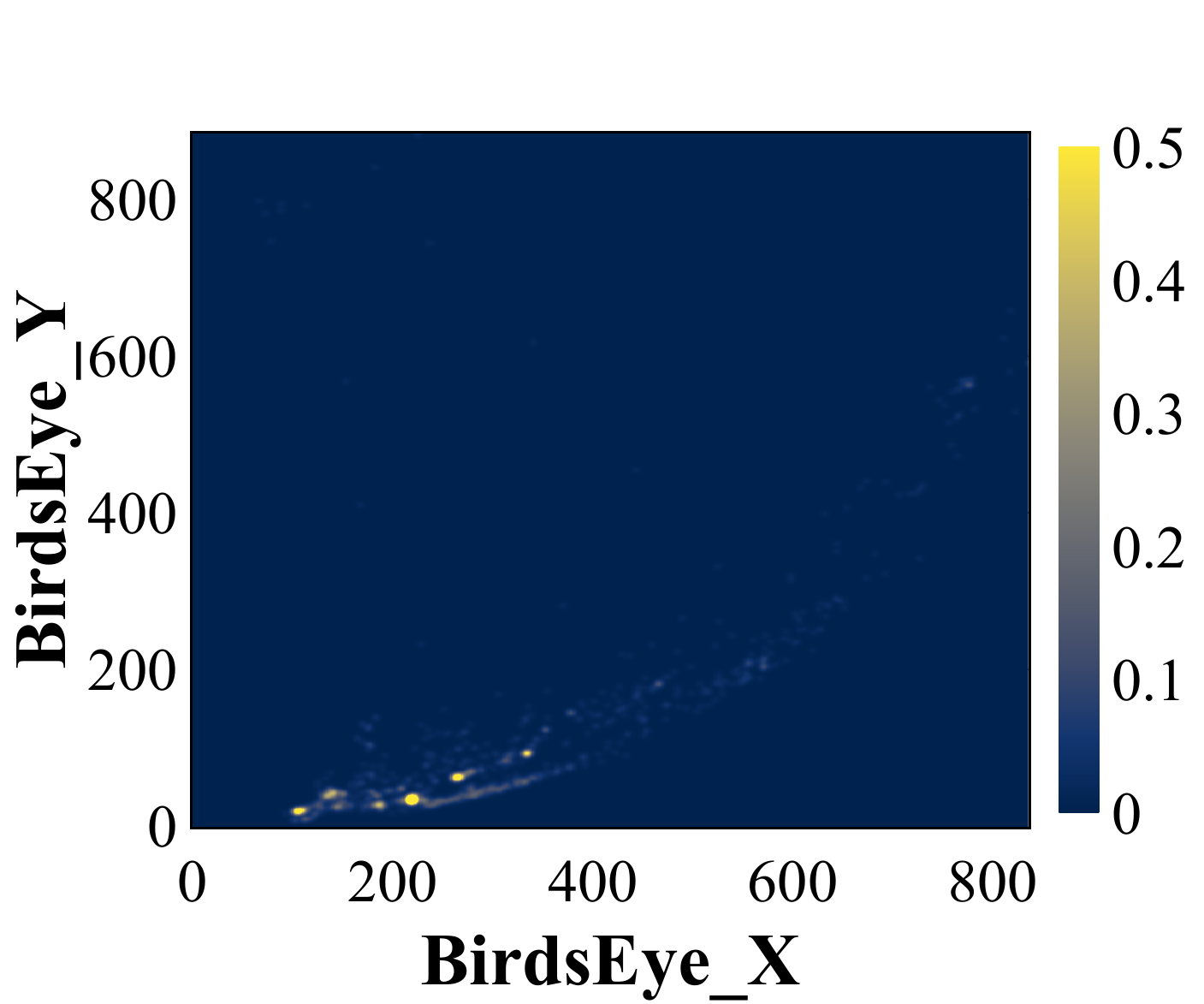}
    }
    \caption{Heatmap representations from two cameras during weekdays and weekends. Graph (a) and (c) represent the heatmaps during weekdays, while Graph (b) and (d) show congestion in the same areas during weekends.}
    \label{fig:Heatmap}
\end{figure}

Heat maps visually represents data distribution, ann are useful for monitoring congestion and spatial usage. The system generates 2D heatmaps from Bird's Eye View coordinates, applying Gaussian smoothing for clarity. Fig. \ref{fig:Heatmap} presents heatmaps for different 24-hour intervals. Graph (a) and (c) show weekday pedestrian movement, while Graph (b) and (d) represent the weekend congestion at the same areas. Comparing the density patterns during weekdays and weekends  reflect differences in area usage.

Algorithm \ref{alg:heatmap} is developed to generate heatmaps based on the data from cameras. We used the calculated Bird's Eye View coordinates to generate heat maps. We processed data for each of the eight cameras, filtering records for every 24-hour window within the specified date range. 

\begin{algorithm}[tb]
   \caption{Heatmap Generation}
   \label{alg:heatmap}
\begin{algorithmic}[1]
   \STATE \textbf{Input:} data frame $df$, start\_date, end\_date
   \STATE Compute date\_range from start\_date to end\_date
   \FOR{each date in date\_range}
       \FOR{$camera\_id$ from 1 to 8}
           \STATE Filter $df$ for current date and $camera\_id$
           \IF{filtered data is not empty}
               \STATE Determine $max\_x$ and $max\_y$ from filtered data
               \STATE Initialize heatmap array of size $(max\_y + 1, max\_x + 1)$
               \FOR{each row in filtered data}
                   \STATE Increment heatmap value at position (Bird's Eye $Y$, Bird's Eye $X$)
               \ENDFOR
           \ENDIF
       \ENDFOR
   \ENDFOR
\end{algorithmic}
\end{algorithm}

The initial step involves defining a date range between a given \texttt{start date} and an \texttt{end date}. The algorithm then iterates over this date range and, for each date, processes data from each of the eight cameras. This systematic approach ensures comprehensive coverage, capturing the spatial dynamics of the monitored area over time.

For each date and camera combination, the dataset, \( df \), is filtered to include only records that fall within the current date and are associated with the current camera. Once filtered, the algorithm checks if there's any data available for processing.

If data is available for the given date and camera, the maximum Bird's Eye View coordinates, \( \texttt{max x} \) and \( \texttt{max y} \), are determined. These values define the dimensions of a 2D heatmap array, which is initialized with zeros. As the algorithm iterates over each row of the filtered data, the heatmap values are incremented based on the Bird's Eye View positions of detected objects.  Thus, by the end of this process, we obtain a series of heat maps, each capturing the spatial distribution of activity for a specific camera on a particular day, offering invaluable insights into the daily patterns of area usage and congestion. 
We applied Gaussian smoothing to the original heatmaps to enhance their clarity.

\section{Community Engagement}\label{sec:engagement}

To gauge public perception of our solution, we conducted seven rounds of survey studies in six public places across Charlotte, NC, during July and August 2023. A total of 410 respondents participated, providing insights into their concerns about safety and surveillance. 

The results indicated that respondents were most concerned about safety in parking lots (54\%), public transit (44\%), and entertainment venues (42\%). Additionally, significant concerns were raised about potential biases and discrimination in current passive surveillance technologies, being rated as important or very important to respondents across all demographic groups, highlighting a widespread sensitivity to this issue. \footnote{This study has been previously published, but in adherence to the double-blind review policy of the AAAI conference, the reference will be provided in the camera-ready version upon acceptance. }

Interestingly, vulnerable groups, such as females and people of color, were statistically more likely to agree with the statement, "The presence of video surveillance makes me feel safer." Furthermore, respondents found our AI-driven video solution more beneficial, with over 50\% rating it as very or extremely beneficial, compared to current passive technologies, which 42\% rated as not effective or only moderately effective. Our solution also raised fewer privacy concerns, with 66\% of respondents reporting no privacy concerns or only moderate concerns. These findings suggest a strong preference for AI-driven surveillance solutions over traditional passive technologies, especially in public places.

\section{System Evaluation and Results}\label{sec:Results} 

\subsection {Load Stress Evaluation}
This section evaluated the system's performance under increasing input loads by varying crowd densities from videos. Key metrics such as average latency and throughput were measured as we scaled the system using different numbers of pipeline nodes—one, four, eight, and twelve—to assess scalability and adaptability under increased workloads.

Our performance evaluation methodology employed a systematic testing approach using the CHAD dataset \cite{CHAD}, comprising ten videos with crowd density levels ranging from 0 to 9. Each test video was 150 seconds in duration with a 30-frames-per-second capture rate. To ensure robust statistical analysis and minimize the impact of system initialization and termination effects, we focused our analysis on the central 100 batches of each test run, excluding the first 25 batches (warm-up period) and the final 25 batches (cool-down period) from our calculations.

\begin{figure}[t]
    \centering
    \subfloat[]
    {
        \includegraphics[width=0.43\textwidth, trim= 10 0 60 70,clip]{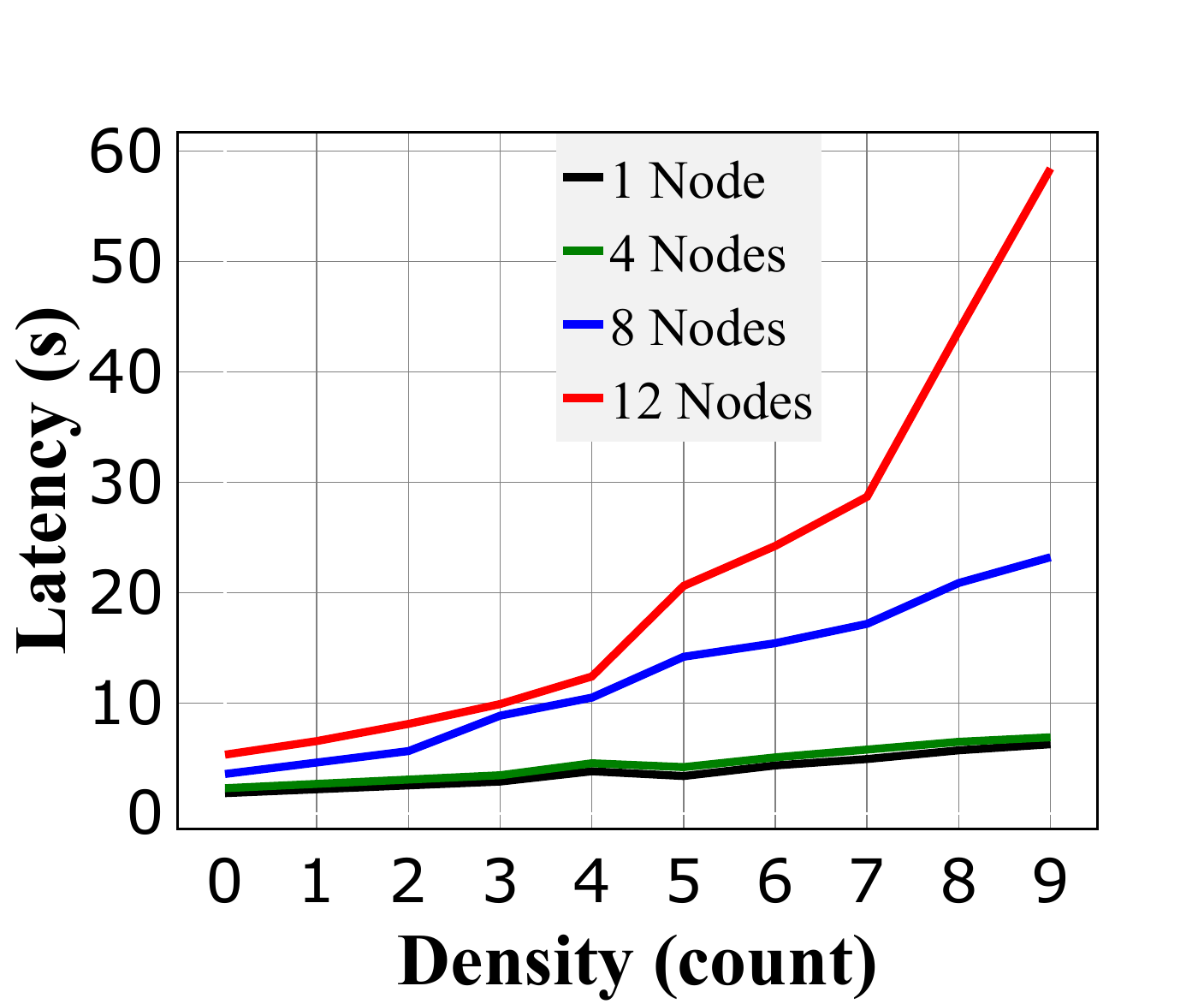}
    }
    \hfill
    \subfloat[]
    {
        \includegraphics[width=0.43\textwidth, trim= 10 0 60 70,clip]{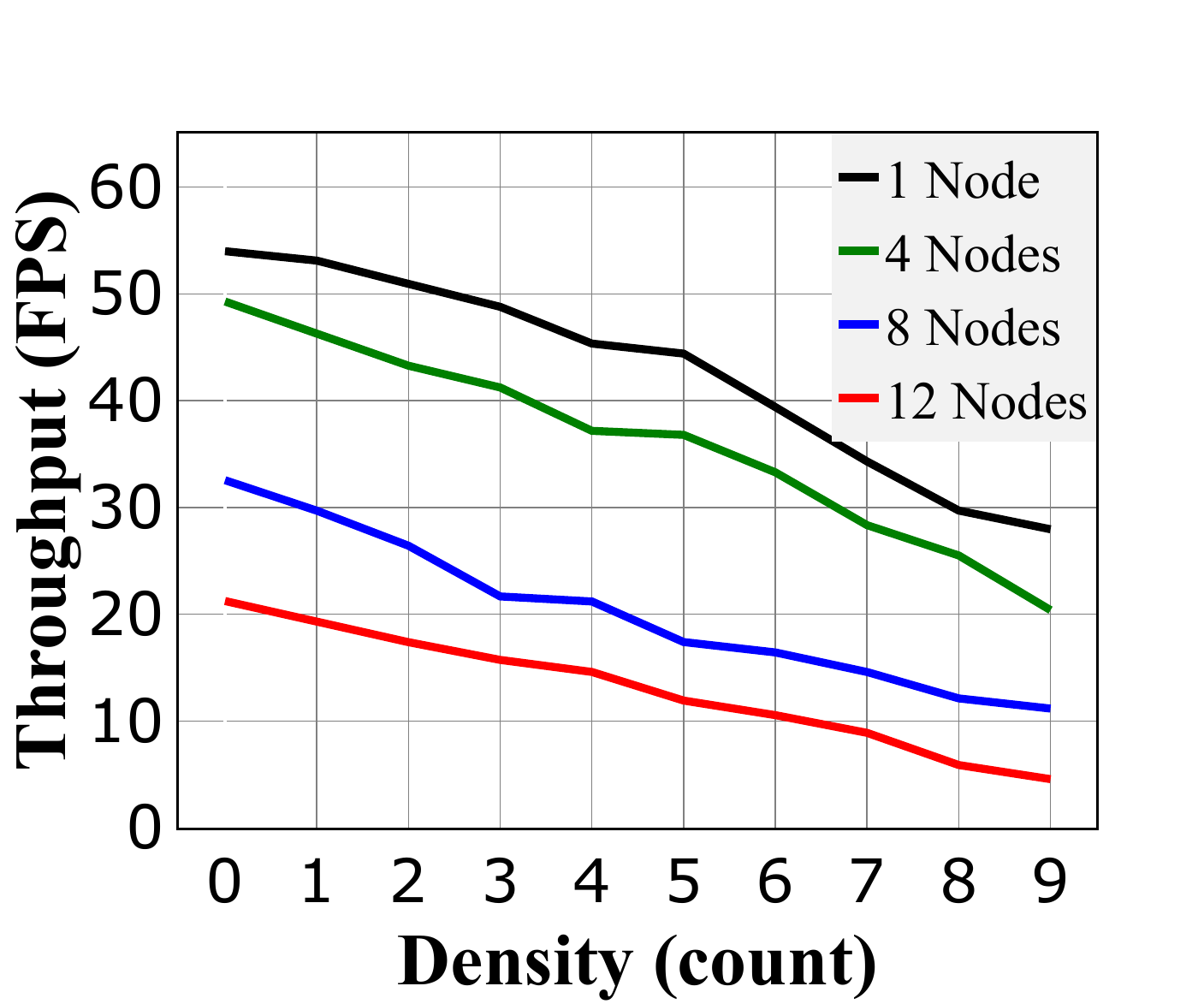}
    }
    \caption{Latency and Throughput trends with respect to crowd densities across different number of nodes running in parall. Graph (a) shows the latency trends, and Graph (b) shows the throughput trends.}
    \label{fig:density}
\end{figure}

Figure \ref{fig:density} illustrates the relationship between system performance metrics and crowd density. The X-axis represents the average number of individuals detected per 30-frame batch across varying input node configurations. System performance exhibited distinct patterns across different node configurations. With one to four camera nodes, the system maintained stable latency below 10 seconds across all density levels. However, the eight-node configurations showed significant latency increases exceeding 20 seconds at density levels above 5. Interestingly, the twelve-node configuration demonstrated improved latency at high density levels (8 and 9), particularly when processing approximately 108 individuals.

System throughput exhibited a linear decline correlating with increases in both density and node count, reaching its minimum of 4.56 FPS at density level 9 with 12 nodes. This performance degradation primarily stems from two factors: the computational intensity of the pose estimator \cite{hrnet} processing dense keypoint configurations, and CPU processing limitations. The combination of these factors creates a performance bottleneck that intensifies with increasing node count and crowd density.

\subsection{Real-World Endurance Evaluation}

This evaluation assessed the system's long-term stability using all 16 cameras in the testbed. We conducted extended trials with eight, twelve, and sixteen nodes running continuously for 21 hours, and a week-long test using eight nodes to ensure the system's robustness. This setup aimed to validate the system's performance in realistic, extended operational conditions.

\begin{itemize}
\item \textbf{Eight-Node Setup:} Covers key indoor areas, including entry points, hallways, and common spaces.
\item \textbf{Twelve-Node Setup:} Adds outdoor monitoring of parking lots.
\item \textbf{Sixteen-Node Setup:} Utilizes all cameras for comprehensive coverage.
\end{itemize}

\begin{figure*}[]
        \centering
               \includegraphics[width=1\linewidth, trim= 10 0 10 20,clip]{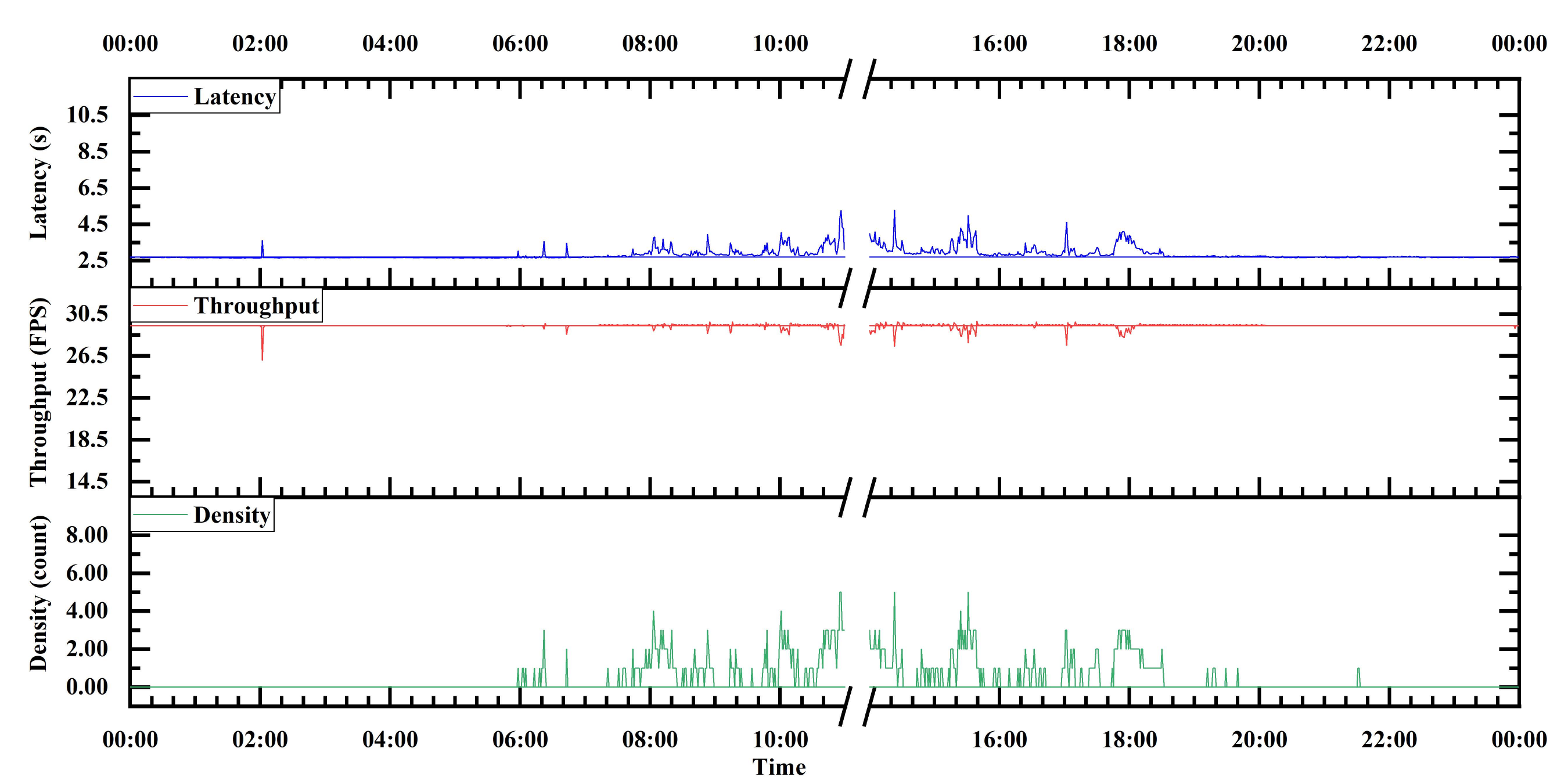}
                    \caption{Latency and throughput trends concerning crowd densities during a 21-hour-length period for 8 camera nodes}
                \label{fig:21_8cam}
\end{figure*}

\begin{figure*}[]
        \centering
               \includegraphics[width=1\linewidth, trim= 10 0 10 20,clip]{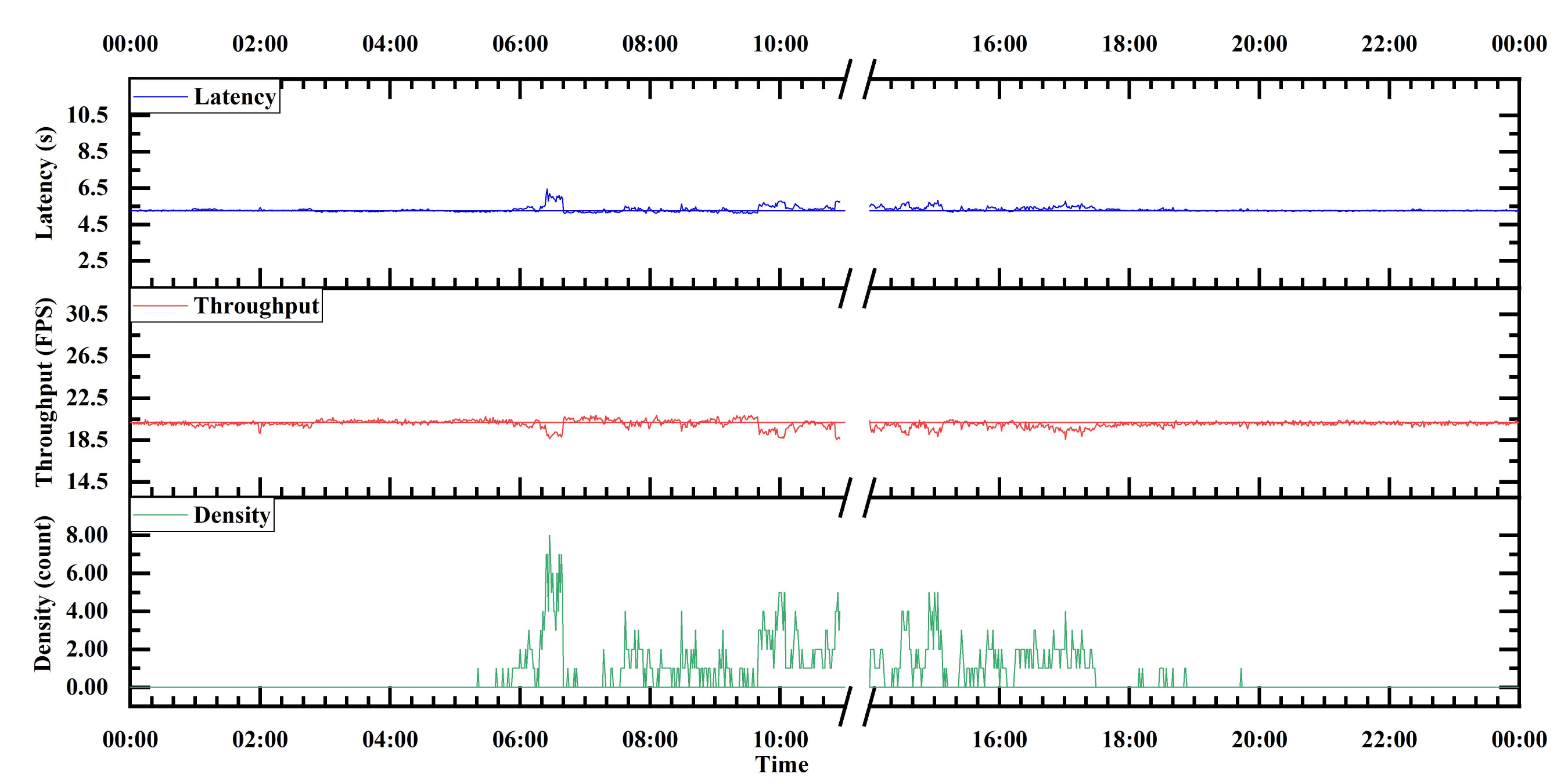}
                    \caption{Latency and throughput trends concerning crowd densities during a 21-hour-length period for 12 camera nodes}
                \label{fig:21_12cam}
\end{figure*}

\begin{figure*}[]
        \centering
               \includegraphics[width=1\linewidth, trim= 10 0 10 20,clip]{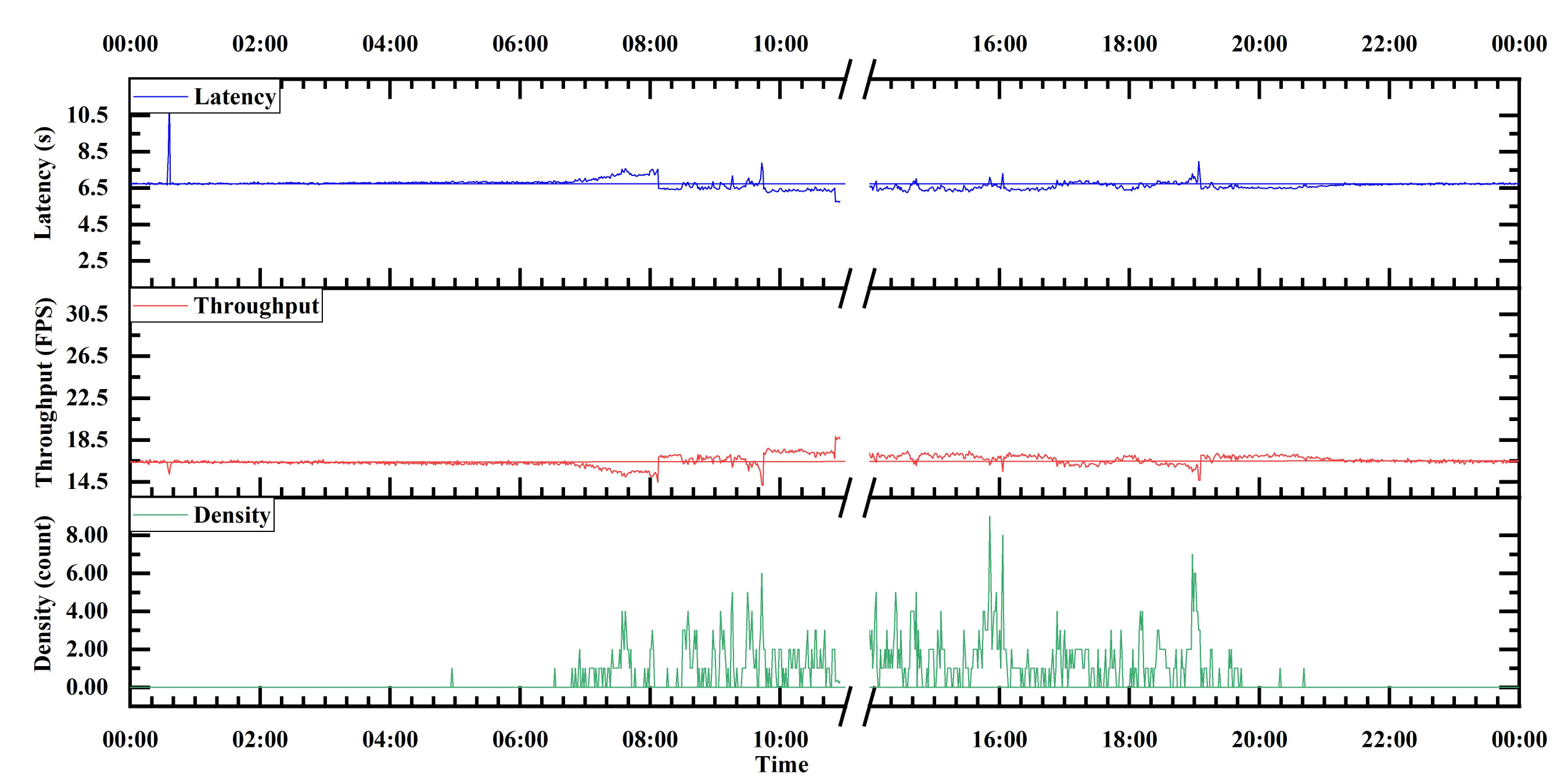}
                    \caption{Latency and throughput trends concerning crowd densities during a 21-hour-length period for 16 camera nodes}
                \label{fig:21_16cam}
\end{figure*}

Figures \ref{fig:21_8cam}, \ref{fig:21_12cam}, and \ref{fig:21_16cam} present comprehensive analyses of system latency and throughput patterns across varying crowd densities during a 21-hour operational period. The evaluation, conducted from 2:00 p.m. to 11:00 a.m. the following day, was tested with three camera node configurations. Our observations revealed a notable decrease in system latency when the server node's database approached 50,000 queries. To maintain optimal performance and ensure data privacy, we implemented an automatic database reset mechanism that activates either upon reaching 50,000 queries or after 24 hours of continuous operation.

The data presented in Figures \ref{fig:21_8cam}, \ref{fig:21_12cam}, and \ref{fig:21_16cam} represent values averaged over 60-second intervals. The 'Density' metric reflects the aggregate number of human detections per camera node during these one-minute sampling windows. The relatively low human traffic observed, particularly from indoor camera nodes, can be attributed to the data collection period coinciding with the summer break period.

\begin{figure*}[]
        \centering
               \includegraphics[width=1\linewidth, trim= 0 0 0 10,clip]{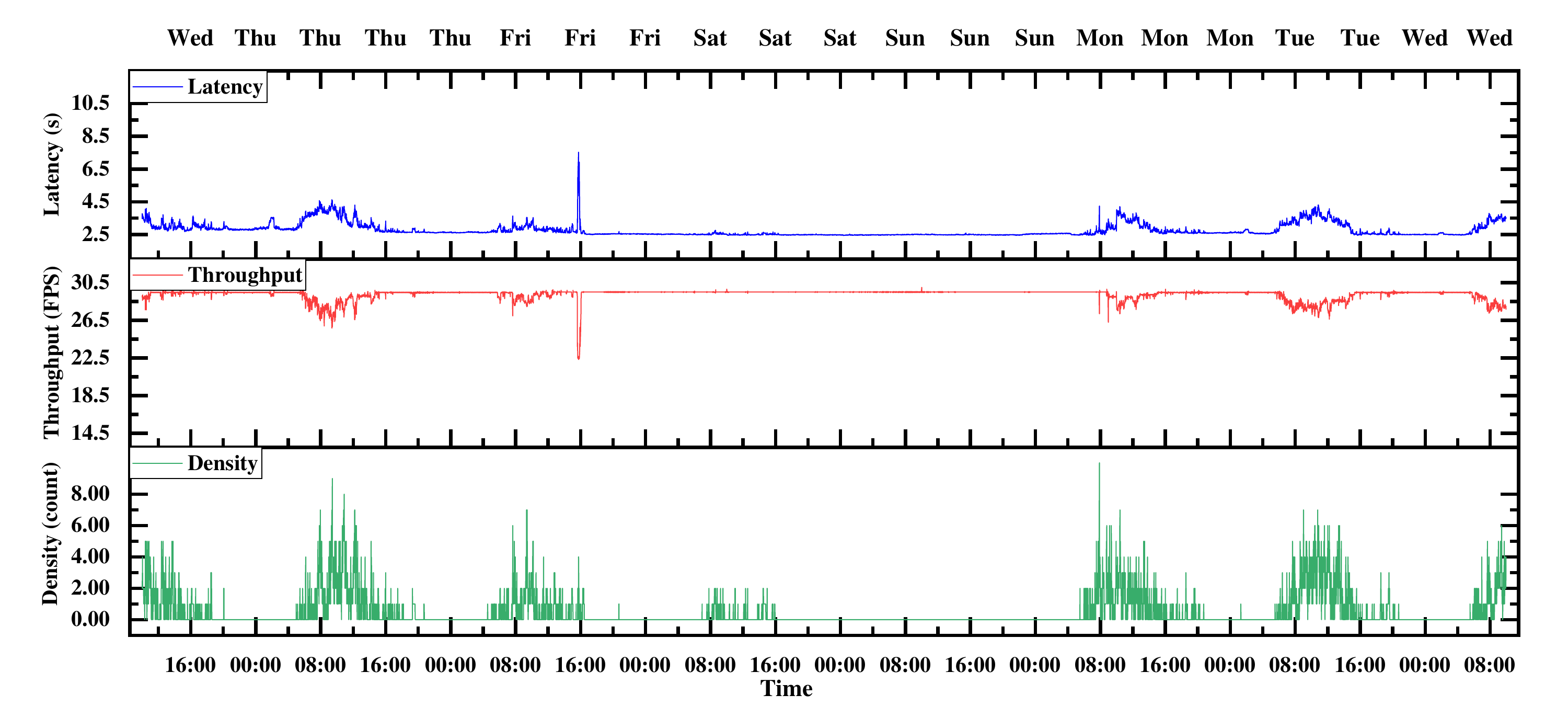}
                    \caption{Latency and throughput trends concerning crowd densities during a week-long period for 8 camera nodes}
                \label{fig:week}
\end{figure*}

Figure \ref{fig:week} illustrates the system's week-long performance using an eight-camera configuration. The results demonstrate consistent correlations between crowd density and performance metrics, with latency ranging from 2.5-7.8 seconds and throughput varying between 22.3-29.8 FPS. These findings confirm our expected inverse relationship between crowd density and system performance, while demonstrating that video duration has minimal impact on overall system performance.

Within the dataset, occasional performance spikes were observed, potentially attributable to environmental network fluctuations from IP camera streams or periods of peak CPU memory utilization. However, these variations remained within acceptable operational parameters, demonstrating the system's resilience under diverse conditions.
The extended 21-hour evaluation across different camera configurations revealed consistent performance patterns:

\begin{itemize}
\item \textbf{8 nodes:} latency of 2.6-4.8s with throughput of 28.5-26.5 FPS
\item \textbf{12 nodes:} latency of 5.3-6.5s with throughput of 20.5-18 FPS
\item \textbf{16 nodes:} latency of 6.7-10.5s with throughput of 16.5-14.5 FPS
\end{itemize}

The system demonstrated consistent performance under diverse operational conditions, validating its robustness and adaptability.

\subsection{Physical-Cyber-Physical Evaluation (Anomaly detection)}
\label{sec:anomaly}

\begin{figure*}[]
        \centering
               \includegraphics[width=1\linewidth, trim= 0 190 0 0,clip]{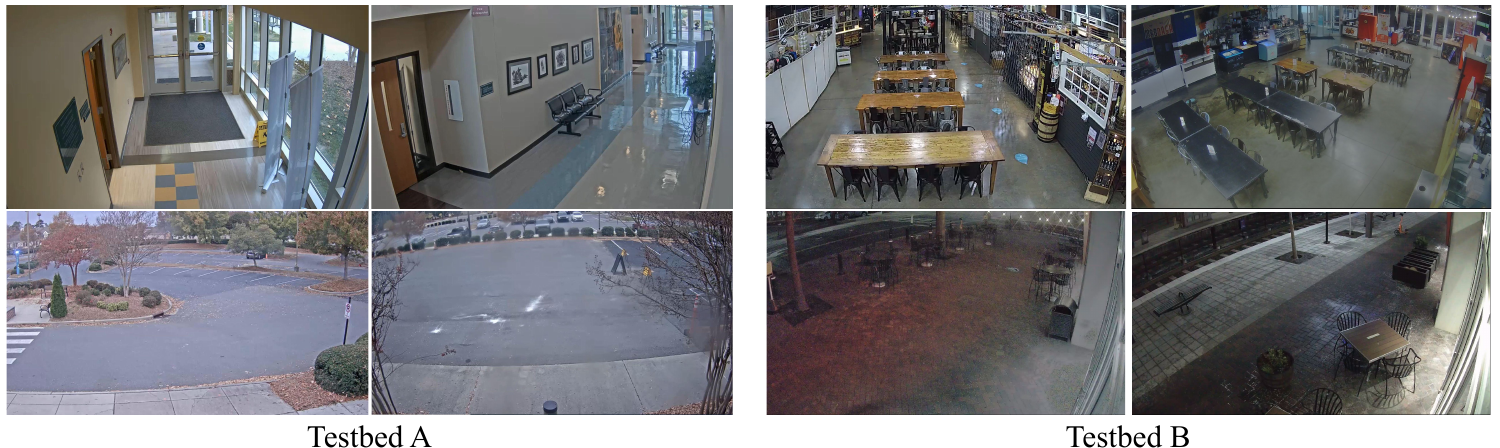}
                    \caption{Example footages of the cameras in different locations}
                \label{fig:footage}
\end{figure*}

This section evaluates the system’s Physical-Cyber-Physical (PCP) latency, representing the end-to-end time from when an anomaly appears to when the end user receives a notification. To showcase the versatility and robustness of our AI-driven surveillance system, we established Testbed B with 12 CCTV cameras—3 outdoor and 9 indoor—at a busy market shop, featuring a 32-core CPU, 512 GB RAM, and two 48 GB VRAM GPUs. This setup complements Testbed A, providing diverse environmental conditions and hardware configurations for testing anomaly detection. For a comparative baseline, we deployed the same pipeline on AWS cloud, using an EC2 p2.xlarge instance with a single GPU to process video streams from IoT IP cameras over Real-Time Streaming Protocol. Metadata is stored in Amazon RDS, queried via Athena, and notifications are sent through SNS, mirroring the edge architecture for a direct performance comparison. Fig. \ref{fig:footage} illustrates example camera views from two testbeds without any human presence.

By comparing the edge-based setups with the cloud deployment, we specifically examined end-to-end latency differences. In the cloud setup, the EC2 instance consolidates local and global processing nodes, with Amazon services enabling real-time analytics and alerts. This configuration highlights how offloading inference to the cloud impacts response times, with results from the p2.xlarge instance serving as a reference point. While alternative AWS configurations could yield different latencies and costs, this setup provides valuable insights into the trade-offs of cloud processing, aiding in the optimization of surveillance system design based on operational needs.

We assessed both object and behavioral anomaly detection \cite{gepc} using four, eight, twelve, and sixteen AI pipeline nodes reading input frames from the same camera node, conducting fifty experiments per anomaly type and collecting 700 data samples.

PCP latency measures the time from detecting anomalies, such as high-priority objects or dangerous behaviors like fighting, to alerting the end user. As mentioned in Sec \ref{sec:System-Features}, the object detector can detect object-of-interest classes as object anomaly and behavior anomaly detector is trained on normal movement sequences and evaluates incoming sequences to see whether the movement is normal or not. For evaluation, we used predefined test cases: prohibited items were placed in predetermined locations, and staged physical confrontations were enacted by trained, authorized participants within the detection range of the camera under examination. These controlled scenarios ensured consistent triggering of the system’s anomaly detection mechanisms, enabling accurate measurement of end-to-end alert latency.

Various strategies were implemented to reflect the system’s versatility across the two testbeds. For testbed A, although the number of camera nodes was increased, multiple streams were duplicated from the same camera, ensuring that crowd density remained controlled. By contrast, testbed B, employed different camera streams onsite when increasing the node count, resulting in an uncontrolled number of individuals entering the field of view, thereby simulating real-world conditions.

About three-second transmission delay from cameras to AI nodes was observed for testbed A, slightly higher than other IP systems, due to its unique network setup. This delay is included in the overall PCP measurements, though manual recording of latency data may introduce minor discrepancies.

Table \ref{tab:PCP detail} provides more detailed insight about P-C-P evaluation results. This evaluation demonstrates the system's capability to enhance safety and operational efficiency in real-world scenarios. The ability to notify end users within approximately 20 seconds enables transforming the existing CCTV systems from passive video cameras to proactive systems that can notify the stakeholders about the events that are happening.

The results indicate that increasing the number of cameras significantly impacts PCP latency. For example, object anomaly latency increased from 4.7 seconds with four cameras to 19.68 seconds with sixteen, while behavioral anomaly latency rose from 9.34 seconds to 26.76 seconds in testbed A. This increase is attributed to higher system traffic, multitasking overhead, and the overall processing capacity. Tests at Testbed B confirmed these trends, illustrating how a more dynamic environment and increased workload can further affect end-to-end alert times. In contrast, the cloud-based system offers significant advantages in terms of performance stability. Unlike the edge system, which is constrained by its local processing capabilities and experiences substantial latency increases as more cameras are added, the cloud infrastructure can dynamically allocate additional processing instances to handle the increased workload from more cameras. However, the result of 21.38 seconds for object anomaly detection and 26.5 seconds for behavior anomaly detection is not competitive with the edge based system in the experiment.

\begin{table}[]
\centering
\caption{Statistical PCP latency data for object and behavior anomaly for different number of cameras in different environment}
\label{tab:PCP detail}
\small 
\setlength\tabcolsep{3pt}
\resizebox{\columnwidth}{!}{%
\begin{tabular}{c|c|cccc|cccc|c}
\rowcolor{DarkGray}
 &  & \multicolumn{4}{c|}{Location A (Controlled)} & \multicolumn{4}{c|}{Location B (Open)} & Full Cloud\\
\rowcolor{Gray}
 & Nodes & \multicolumn{1}{c|}{Mean (s)} & \multicolumn{1}{c|}{Min (s)} & \multicolumn{1}{c|}{Max (s)} & \begin{tabular}[c]{@{}c@{}}Standard \\ deviation\end{tabular} & \multicolumn{1}{c|}{Mean (s)} & \multicolumn{1}{c|}{Min (s)} & \multicolumn{1}{c|}{Max (s)} & \begin{tabular}[c]{@{}c@{}}Standard \\ deviation\end{tabular} & Mean(s) \\ \hline
\multirow{4}{*}{\begin{tabular}[c]{@{}c@{}}Object\\ Anomaly\end{tabular}} & 4 & \multicolumn{1}{c|}{4.7} & \multicolumn{1}{c|}{3.06} & \multicolumn{1}{c|}{6.4} & 0.55 & \multicolumn{1}{c|}{3.372} & \multicolumn{1}{c|}{2.29} & \multicolumn{1}{c|}{5.01} & 0.64 & \multirow{4}{*}{21.38} \\
 & 8 & \multicolumn{1}{c|}{10.99} & \multicolumn{1}{c|}{9.18} & \multicolumn{1}{c|}{12.55} & 0.85 & \multicolumn{1}{c|}{8.924} & \multicolumn{1}{c|}{7.46} & \multicolumn{1}{c|}{10.61} & 0.71 &  \\
 & 12 & \multicolumn{1}{c|}{15.76} & \multicolumn{1}{c|}{14.05} & \multicolumn{1}{c|}{18.31} & 1.05 & \multicolumn{1}{c|}{13.037} & \multicolumn{1}{c|}{10.28} & \multicolumn{1}{c|}{16.93} & 1.27 &  \\
 & 16 & \multicolumn{1}{c|}{19.68} & \multicolumn{1}{c|}{17.67} & \multicolumn{1}{c|}{21.6} & 1.13 & \multicolumn{1}{c|}{-} & \multicolumn{1}{c|}{-} & \multicolumn{1}{c|}{-} & - &  \\ \hline
\multirow{4}{*}{\begin{tabular}[c]{@{}c@{}}Behavior\\ Anomaly\end{tabular}} & 4 & \multicolumn{1}{c|}{9.34} & \multicolumn{1}{c|}{7.81} & \multicolumn{1}{c|}{10.41} & 0.65 & \multicolumn{1}{c|}{6.09} & \multicolumn{1}{c|}{4.78} & \multicolumn{1}{c|}{7.37} & 0.6 & \multirow{4}{*}{26.5} \\
 & 8 & \multicolumn{1}{c|}{14.53} & \multicolumn{1}{c|}{13.23} & \multicolumn{1}{c|}{16.46} & 0.71 & \multicolumn{1}{c|}{11.62} & \multicolumn{1}{c|}{10.75} & \multicolumn{1}{c|}{13.3} & 0.62 &  \\
 & 12 & \multicolumn{1}{c|}{18.45} & \multicolumn{1}{c|}{16.32} & \multicolumn{1}{c|}{20.35} & 0.97 & \multicolumn{1}{c|}{17.68} & \multicolumn{1}{c|}{15.42} & \multicolumn{1}{c|}{20.55} & 1.06 &  \\
 & 16 & \multicolumn{1}{c|}{26.76} & \multicolumn{1}{c|}{24.23} & \multicolumn{1}{c|}{29.54} & 1.52 & \multicolumn{1}{c|}{-} & \multicolumn{1}{c|}{-} & \multicolumn{1}{c|}{-} & - & 
\end{tabular}%
}
\end{table}

Overall, these real-world evaluations allow research into how key parameters—such as the number of cameras and crowd density—affect system performance metrics like latency and throughput. Understanding these dynamics in different public safety scenarios, especially in cases of emergency, is crucial for informed decision-making and timely action. For instance, in retail environments and transportation hubs, the system can provide valuable insights into foot traffic patterns and crowd density fluctuations, facilitating both operational optimization and emergency response. \cite{hosseini2024understanding}

\section{Conclusion}\label{sec:Conclusion}
This paper presented a comprehensive evaluation of an AI-based real-time video solution, integrated with existing CCTV infrastructures, to enhance situational awareness through object and behavioral anomaly detection. The system was tested across various configurations, including up to sixteen camera nodes, to assess its performance under real-world conditions.

Our evaluation highlighted the system's ability to maintain operational efficiency with increasing node counts, despite a corresponding rise in PCP latency. Object and behavioral anomaly detection latencies ranged from 4.7 seconds with four cameras to 26.76 seconds with sixteen, reflecting the system's scalability and robustness. 
Through extensive testing, the system demonstrated its capability to deliver timely notifications to end users, which is crucial for managing anomalies in public spaces. This work underscores the importance of real-world evaluations in optimizing AI-driven video solutions, ensuring they meet the demands of dynamic environments and contribute to enhanced public safety and operational efficiency.

\bibliography{main}

\end{document}